\documentclass[10pt,twocolumn,letterpaper]{article}

\usepackage[accsupp]{axessibility}

\usepackage{wacv}              % To produce the CAMERA-READY version
\usepackage{graphicx}
\usepackage{amsmath}
\usepackage{amssymb}
\usepackage{booktabs}
\usepackage{color, colortbl}
\usepackage{makecell}
	
\definecolor{Gray}{gray}{0.9}
\definecolor{LightCyan}{rgb}{0.88,1,1}

\usepackage[pagebackref,breaklinks,colorlinks]{hyperref}
\usepackage[capitalize]{cleveref}
\crefname{section}{Sec.}{Secs.}
\Crefname{section}{Section}{Sections}
\Crefname{table}{Table}{Tables}
\crefname{table}{Tab.}{Tabs.}

\def\wacvPaperID{2607} % *** Enter the WACV Paper ID here
\def\confName{WACV}
\def\confYear{2024}

\usepackage{subdef}

\begin{document}

\title{\fontsize{16}{15}\selectfont{\Textron : Weakly Supervised Multilingual Text Detection through Data Programming}}

\author{
    % \small{Dhruv Kudale} \\
    % % \small{\texttt{dhruvk@cse.iitb.ac.in}} \\
    % % \small{CSE Department, IIT Bombay, India}
    % \and
    % \small{Badri Vishal Kasuba} \\
    % % \small{\texttt{badrivishalk@cse.iitb.ac.in}} \\
    % % \small{CSE Department, IIT Bombay, India}
    % \and
    % \small{Venkatapathy Subramanian} \\
    % % \small{\texttt{venkatapathy@cse.iitb.ac.in}} \\
    % % \small{CSE Department, IIT Bombay, India}
    % \and
    % \small{Ganesh Ramakrishnan} \\
    % % \small{\texttt{ganesh@cse.iitb.ac.in}} \\
    % % \small{CSE Department, IIT Bombay, India}
    % \and
    % \small{Parag Chaudhuri} 
    % % \small{\texttt{ paragc@cse.iitb.ac.in}} \\
    % % \small{CSE Department, IIT Bombay, India}
    {\fontsize{11}{10}\selectfont Dhruv Kudale, Badri Vishal Kasuba, Venkatapathy Subramanian, Parag Chaudhuri, Ganesh Ramakrishnan, } 
    \and
    \small{\texttt{ \{dhruvk, badrivishalk, venkatapathy, paragc, ganesh\}@cse.iitb.ac.in}} \\
     \small{Department of Computer Science and Engineering } \\
     \small{IIT Bombay, India} \\
}
\maketitle

%%%%%%%%% ABSTRACT
\begin{abstract}
Several recent deep learning (DL) based techniques perform considerably well on image-based multilingual text detection. However, their performance relies heavily on the availability and quality of training data. There are numerous types of page-level document images consisting of information in several modalities, languages, fonts, and layouts. This makes text detection a challenging problem in the field of computer vision (CV), especially for low-resource or handwritten languages. Furthermore, there is a scarcity of word-level labeled data for text detection, especially for multilingual settings and Indian scripts that incorporate both printed and handwritten text. Conventionally, Indian script text detection requires training a DL model on plenty of labeled data, but to the best of our knowledge, no relevant datasets are available. Manual annotation of such data requires a lot of time, effort, and expertise. In order to solve this problem, we propose \Textron, a {\em Data Programming-based approach}, where users can plug various text detection methods into a weak supervision-based learning framework. One can view this approach to multilingual text detection as an ensemble of different CV-based techniques and DL approaches. $\Textron_{ }$ can leverage the predictions of DL models pre-trained on a significant amount of language data in conjunction with CV-based methods to improve text detection in other languages. We demonstrate that $\Textron_{ }$ can improve the detection performance for documents written in Indian languages, despite the absence of corresponding labeled data. Further, through extensive experimentation, we show improvement brought about by our approach over the current State-of-the-art (SOTA) models, especially for handwritten Devanagari text. Code and dataset has been made available at \url{https://github.com/IITB-LEAP-OCR/TEXTRON}

\end{abstract}

%%%%%%%%% BODY TEXT
\section{Introduction}
\label{sec:intro}
\begin{figure*}
     \centering
     \begin{subfigure}[b]{0.3\textwidth}
        \centering
        \includegraphics[scale=0.08]{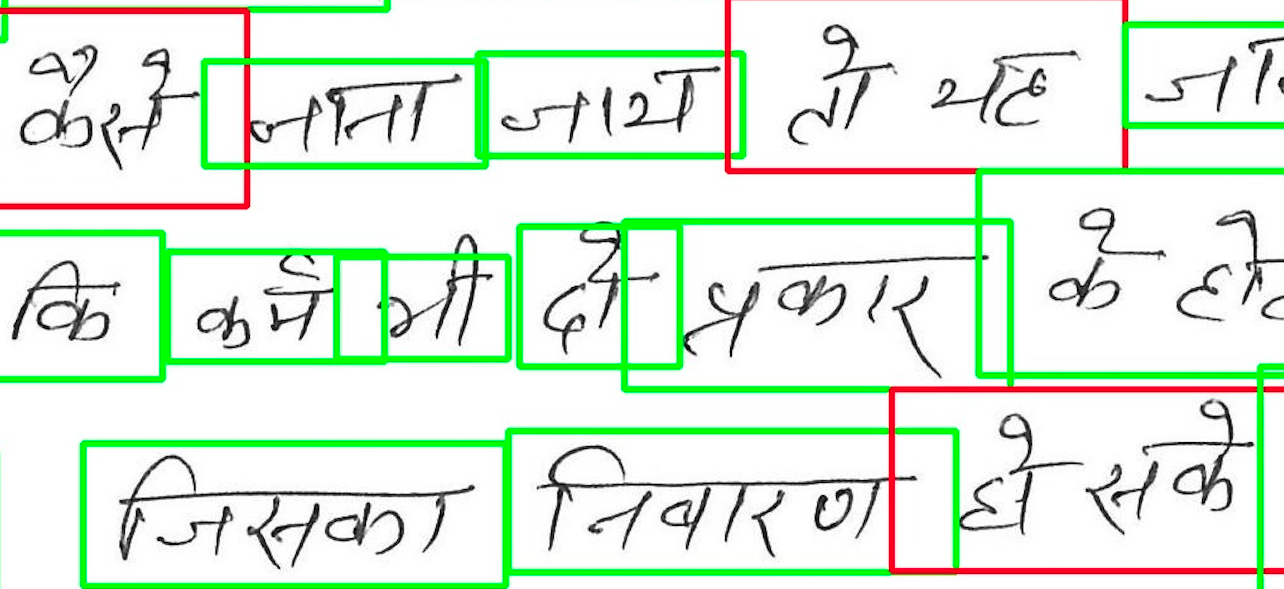}
        \hspace{1mm}
        \caption{DL-Based Method} 
        \label{fig:mot1}
     \end{subfigure}
     \hfill
     \begin{subfigure}[b]{0.3\textwidth}
        \centering
        \includegraphics[scale=0.08]{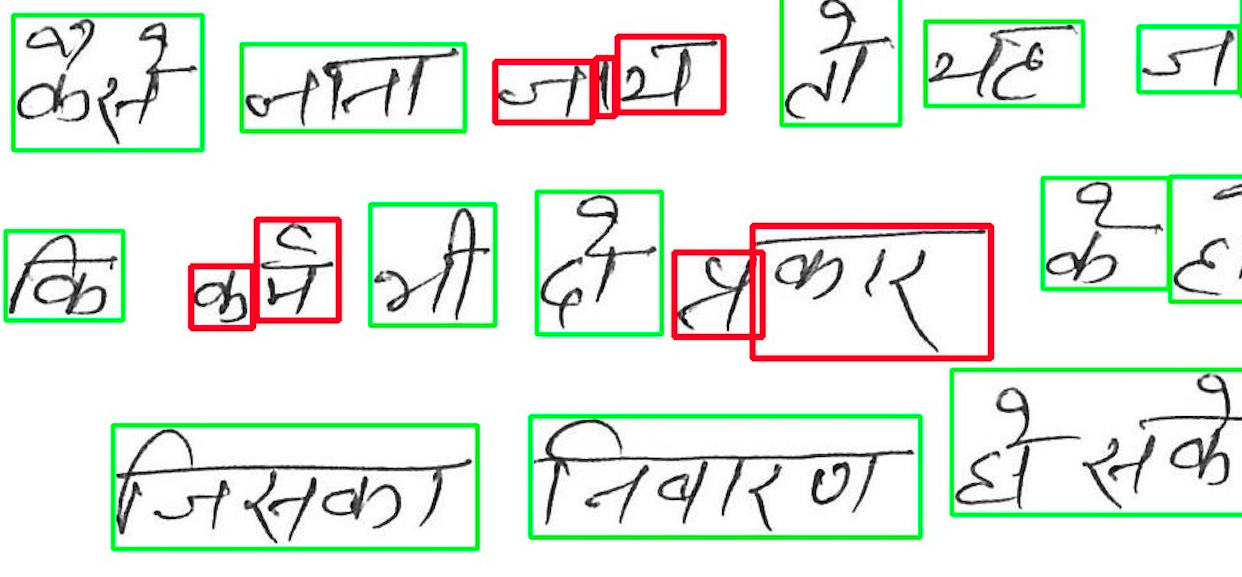}
        \hspace{1mm}
        \caption{CV-based method} 
        \label{fig:mot2}
     \end{subfigure}
     \hfill
     \begin{subfigure}[b]{0.3\textwidth}
        \centering
        \includegraphics[scale=0.08]{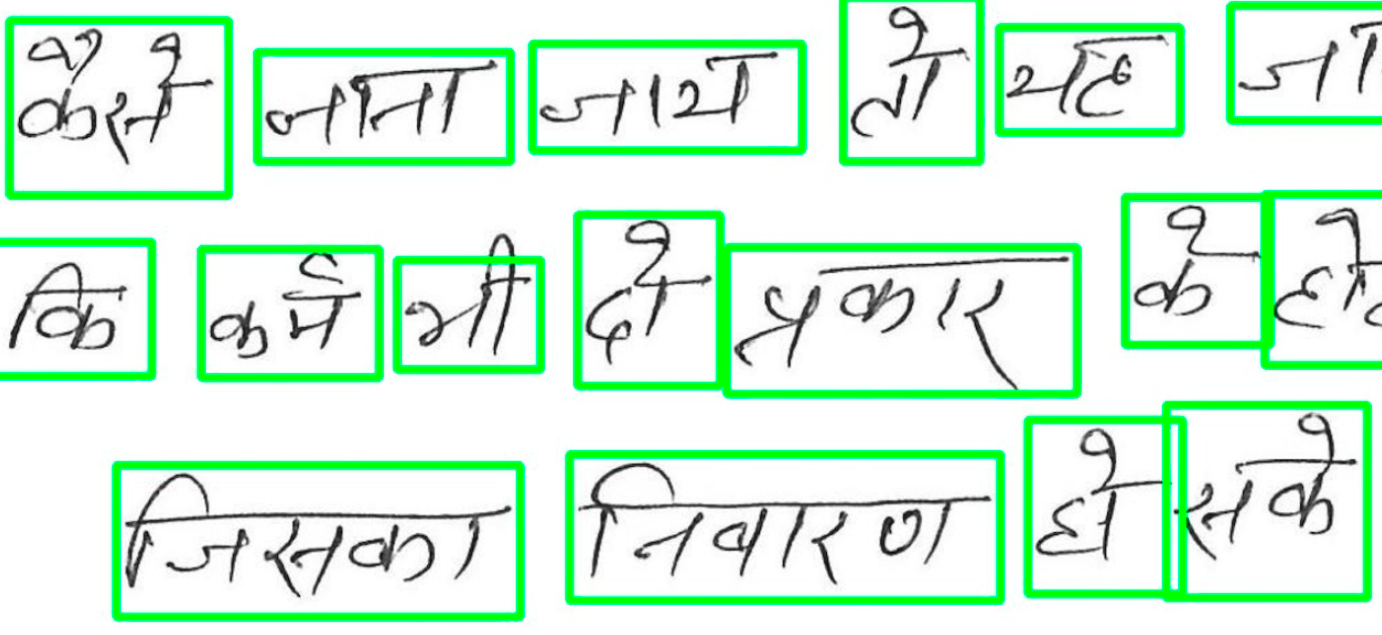}
        \hspace{1mm}
        \caption{Using \Textron} 
        \label{fig:mot3}
     \end{subfigure}
        \caption{A comparative overview of different text detection approaches. The green boxes in subsequent figures highlight true positives while the red ones represent incorrect detections that either engulf more than two words in a single bounding box or fragment a single word into multiple boxes.}
        \label{fig:comparison}
\end{figure*}
Text detection aims to localize the textual information present in images. Text detection from documents is an essential and one of the first steps in Optical Character Recognition (OCR). With Deep Learning(DL) methods replacing conventional approaches in several fields, text detection also benefits from such DL techniques including various convolutional neural networks (CNNs)~\cite{delakis2008text}, region-based convolutional neural networks (R-CNNs), and fully convolutional networks (FCNs)~\cite{zhang2016multi},  offering their robust solutions. A major limitation of deep learning methods is the need for ample training data, which can be challenging to obtain, especially in resource-limited situations like identifying new languages without readily available datasets. Further, as demonstrated in Figure \ref{fig:mot1}, it is seen that pre-trained DL-based methods (like DBNet \cite{dbnet}) are not able to capture different styles of handwriting or word level demarcation for multilingual scripts (such as those of Indian languages). As opposed to this, certain image processing-based techniques like contour-based techniques \cite{contour} which work on the pixel level, can work well to capture these different writing styles by engulfing concerned characters and nuances as illustrated in Figure \ref{fig:mot2}. A lot of CV-based techniques that work on the principle of edge detection, such as Sobel Filter Usage \cite{sobel} or Canny Filter Application \cite{canny} have prevailed for a few decades. Since these CV-based methods that work on pixel-level classification are more adaptive to different styles and fonts of text, they can be considered insightful in contributing to the effective detection of diverse text. However, unlike DL-based methods, CV-based methods are sensitive to noise and may not be able to unify the complete word in one bounding box as can be seen in Figure \ref{fig:mot2}. This can drastically affect the performance of visually rich or scanned documents. Consequently, we propose integrating the robustness and noise immunity of DL-based methods with the diversity-capturing ability of CV-based techniques, leading to the introduction of \Textron. As shown in Figure \ref{fig:mot3}, \Textron{ } can help accurately identify word-level boxes by combining the strengths of both the aforementioned kinds of approaches (CV-based and DL-based). \Textron{ } can help in the effective aggregation of different text detection methods (weak labels) and can also handle scenarios with scarcity or absence of labeled data. This aggregation of different text detection methods is different from an ensemble approach, because in the latter, there is a need to use two or more models to be fitted on the same labeled data, and then the predictions of each model are combined. Contrary to this, \Textron{ } tries to eliminate dependency on labeled data. In an attempt to incorporate different CV-based techniques, \Textron{ } also works on the document images on a pixel-based classification level that helps in capturing the subtlety of text. Thus, through \Textron{ }, we can bring about effective multilingual text detection which is not very sensitive to noise and at the same time helps capture a rich diversity of text written in different languages and modalities (such as handwritten {\em vs.} printed). In addition to this, it also reduces and often eliminates the need to train new DL models and does not rely on labeled data. We highlight recent work in text detection and the role of weak supervision associated with it in Section \ref{section-relwork}. We also describe our methodology and associated unsupervised experimentation in Sections \ref{sec:methodology} and \ref{sec:experiments} respectively. We further demonstrate the usage of $\Textron_{ }$ and report the relevant results in Section \ref{sec:results}. Finally, we conclude with opening doors to more innovative contributions bringing about seamless multilingual text detection.

%-------------------------------------------------------------------------

\section{Related Works}
Deep learning-based text detection has garnered recent attention. Such DL-based text detection in images can be either bounding box-based or pixel-based. Bounding box-based text detection takes inspiration from Object Detection approaches such as YOLO~\cite{yolo}, SSD~\cite{ssd}, and Faster RCNN~\cite{faster-rcnn}. On the other hand, pixel-based text detection methods include the approaches inspired by Mask RCNN~\cite{mask-rcnn} and FCN \cite{fcn} which include CRAFT \cite{craft}, PSENet~\cite{psenet}, etc. Differential Binarization or DBNet \cite{dbnet} is one of the recently introduced approaches that work at the pixel level followed by a label generation process to retrieve word-level bounding boxes. 
% The network generates a feature map which is further used to determine a probability map in conjunction with a threshold map. Both these maps are used to perform differential binarization. The benefit of this approach is the fact that the binarization is differentiable which could be optimized properly. 
% The threshold map determined by the DBNet-generated feature map consists of an adaptive threshold which is defined by a text border map \cite{textborder}. Thus, 
DBNet is a segmentation-based detection network with a RESNET \cite{resnet} backbone making it faster, more accurate, and lightweight. Other segmentation-based text detection methods include LinkNet~\cite{linknet} which attempts to exploit parameters utilization of neural networks efficiently. Due to the optimization in parameter usage, LinkNet helps to decrease the processing time required for text detection. Unsupervised segmentation approaches have also been employed generically to classify regions of an input image and assign labels to them. One such unsupervised approach~\cite{unsupervised} leverages a pre-trained CNN model to assign labels to pixels. However, the number of unique labels should be sufficiently large to leverage this unsupervised segmentation technique. Computer Vision (CV) based text detection includes a variety of algorithms such as edge detection~\cite{sobel}, connected component analysis \cite{koo2013scene,hu2020text}, Stroke Width Transform (SWT)~\cite{swt}, and contour-based methods \cite{contour}. Edge detection methods include the usage of the Sobel filter \cite{sobel} method to find the corresponding edges of the textual information present in a binarized map of the input document image. Further, the canny~\cite{canny} algorithm to detect regions based on edges can also help perform text detection. Recently, to address the scarcity of labeled data, there has been an emergence of weak-supervision and semi-supervision-based text detection in documents. Text-as-Lines \cite{text-as-lines} proposes a scene text detector based on weakly supervised learning that helps in the annotation process. This method uses coarse line-level pixel-based masks to denote text lines in documents. The usage of coarse masks as opposed to full pixel-level masks makes the process robust and faster to train. WeText \cite{wetext} exploits the paradigm of both weak supervision and semi-supervision to eventually train models to perform scene text detection. It follows a character-based approach to handle multilingual as well as multi-oriented text. Various semi-supervised and unsupervised data programming approaches \cite{snorkel, spear, lfsurvey, abhishek2022spear} have proven to be useful in creating labeled data. In our work, we leverage 
%the functionality of SPEAR \cite{spear}, used for 
the data programming paradigm of weakly supervised learning as in CAGE~\cite{cage} that also provides domain-based quality guides on the labeling functions \cite{maheshwari2022learning}.

%It gives users the provision of semi-supervision using the Joint Learning \cite{jointl} approach as well as weak supervision through Chatterjee {\em et al}~\cite{cage}.
\label{section-relwork}

\section{Our Methodology}
We begin by explaining the notion of labeling functions 
that get used for 
%weak supervision through
data programming. 
\label{sec:methodology}

\subsection{Labeling Functions}
Labeling Functions (LFs) \cite{lfsurvey} are conventionally weak supervision functions that generate noisy labels. In our context, LFs mark regions of document image as \textlabel{} or \nontext{} regions. In our setting, the LFs operate at the pixel level by assigning corresponding labels to each pixel. Following are some of the LFs used in our framework to identify \textlabel{} regions. The output predictions from each of the LF would be a binarized pixel map. Each LF is associated with one of the two classes: \textlabel{} for labeling a textual region in the document and \nontext{} otherwise. While some LFs are based on conventional Computer Vision (CV) techniques, others make use of DL-based models:
\label{lfs}

\begin{figure}
  \begin{subfigure}[b]{0.5\linewidth}
    \centering
    \includegraphics[width=\linewidth]{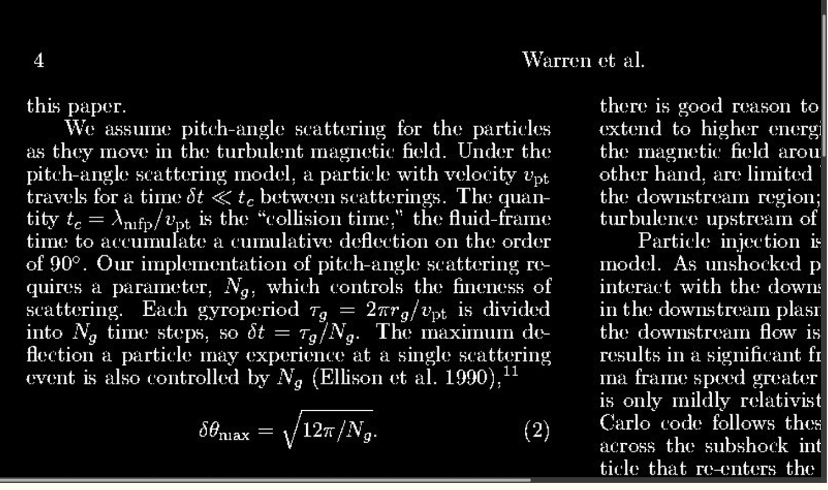} 
    \caption{Pixel-level binarized input} 
  \end{subfigure}%%
  \begin{subfigure}[b]{0.5\linewidth}
    \centering
    \includegraphics[width=\linewidth]{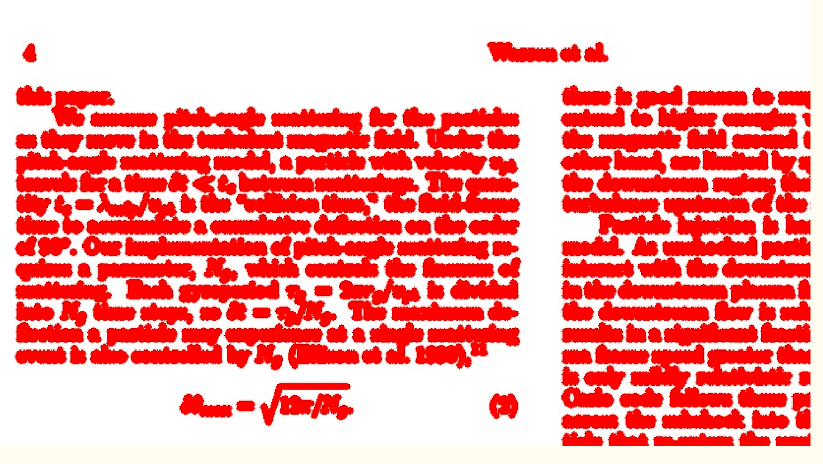} 
    \caption{Finding contours (red)} 
  \end{subfigure} 
  \begin{subfigure}[b]{0.5\linewidth}
    \centering
    \includegraphics[width=\linewidth]{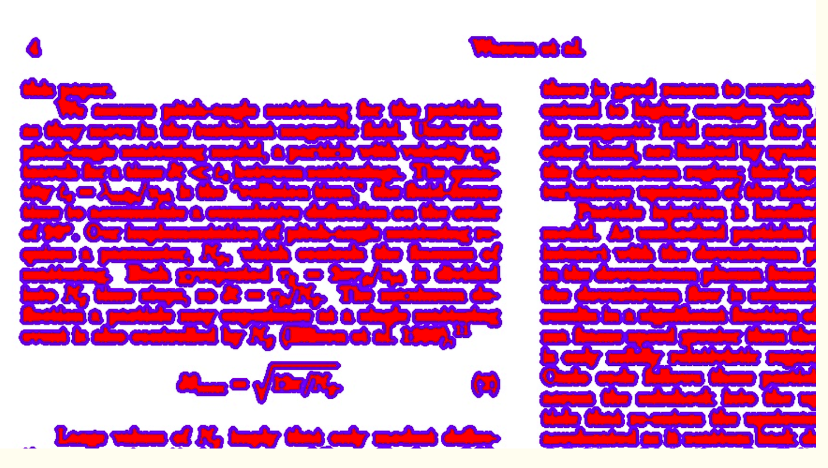} 
    \caption{Drawing contours over the top of red contours} 
  \end{subfigure}%% 
  \begin{subfigure}[b]{0.5\linewidth}
    \centering
    \includegraphics[width=\linewidth]{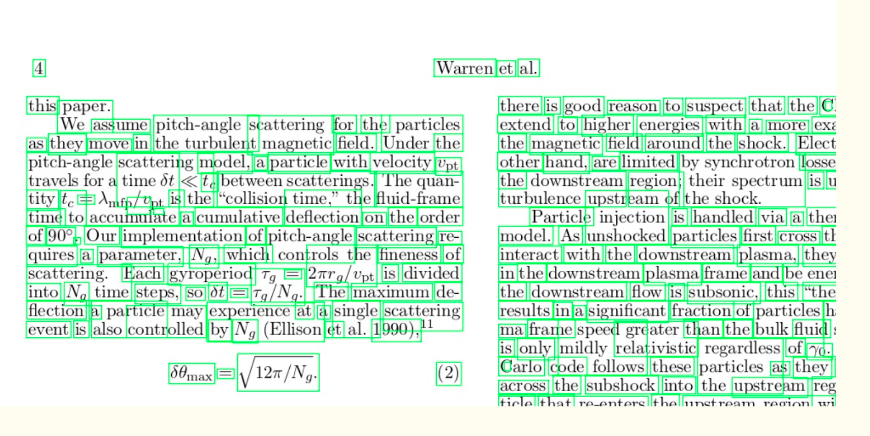} 
    \caption{Get bounding boxes from blue contours} 
    % \vspace{4ex}
  \end{subfigure} 
  \caption{Workflow for Contour-based processing to get word level bounding boxes}
  \label{contourlf} 
\end{figure}

\begin{figure*}
     \centering
     \begin{subfigure}[b]{0.15\textwidth}
        \centering
        \includegraphics[scale=0.03]{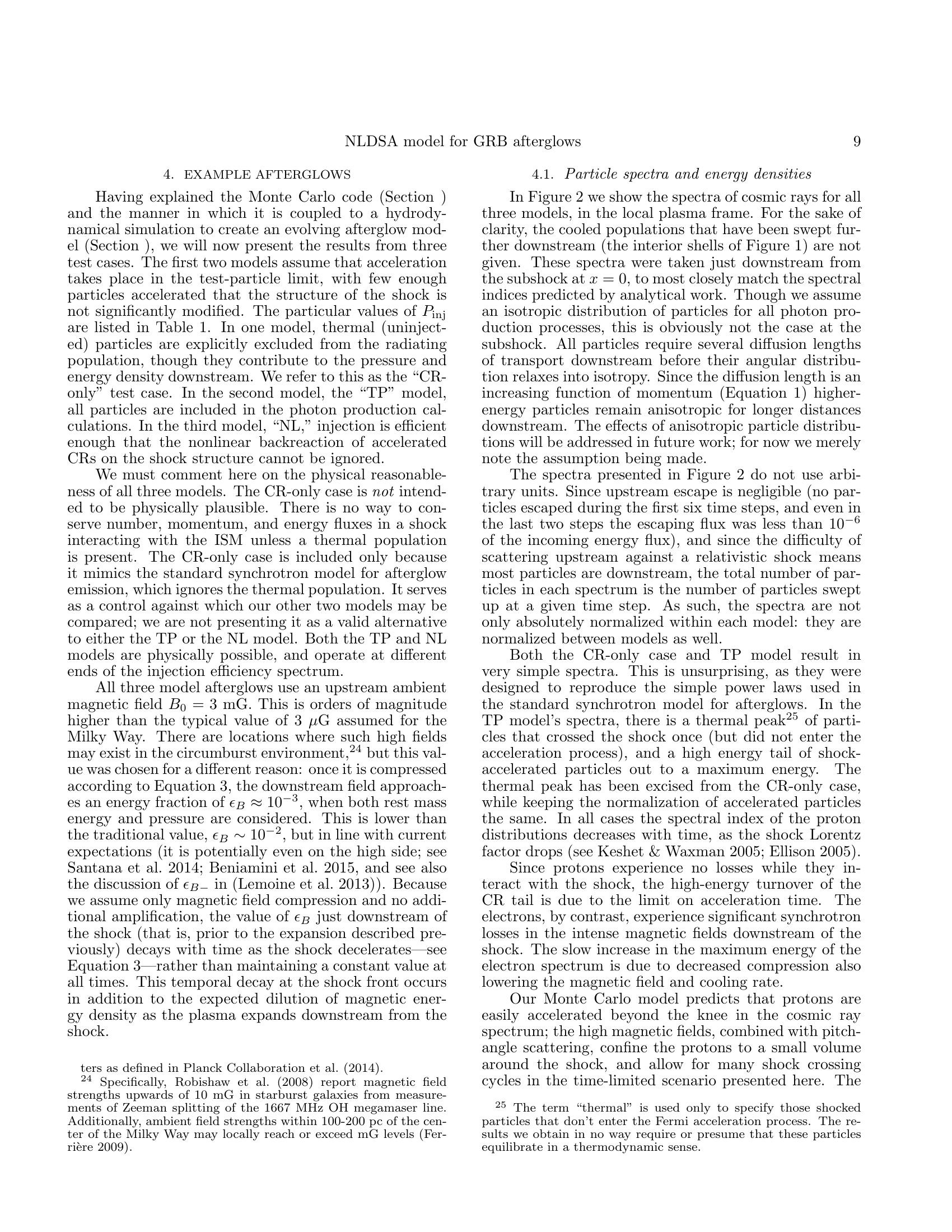}
        \hspace{1mm}
        \caption{Input Image} 
        \label{fig:ex1}
     \end{subfigure}
     \hfill
     \begin{subfigure}[b]{0.15\textwidth}
        \centering
        \includegraphics[scale=0.03]{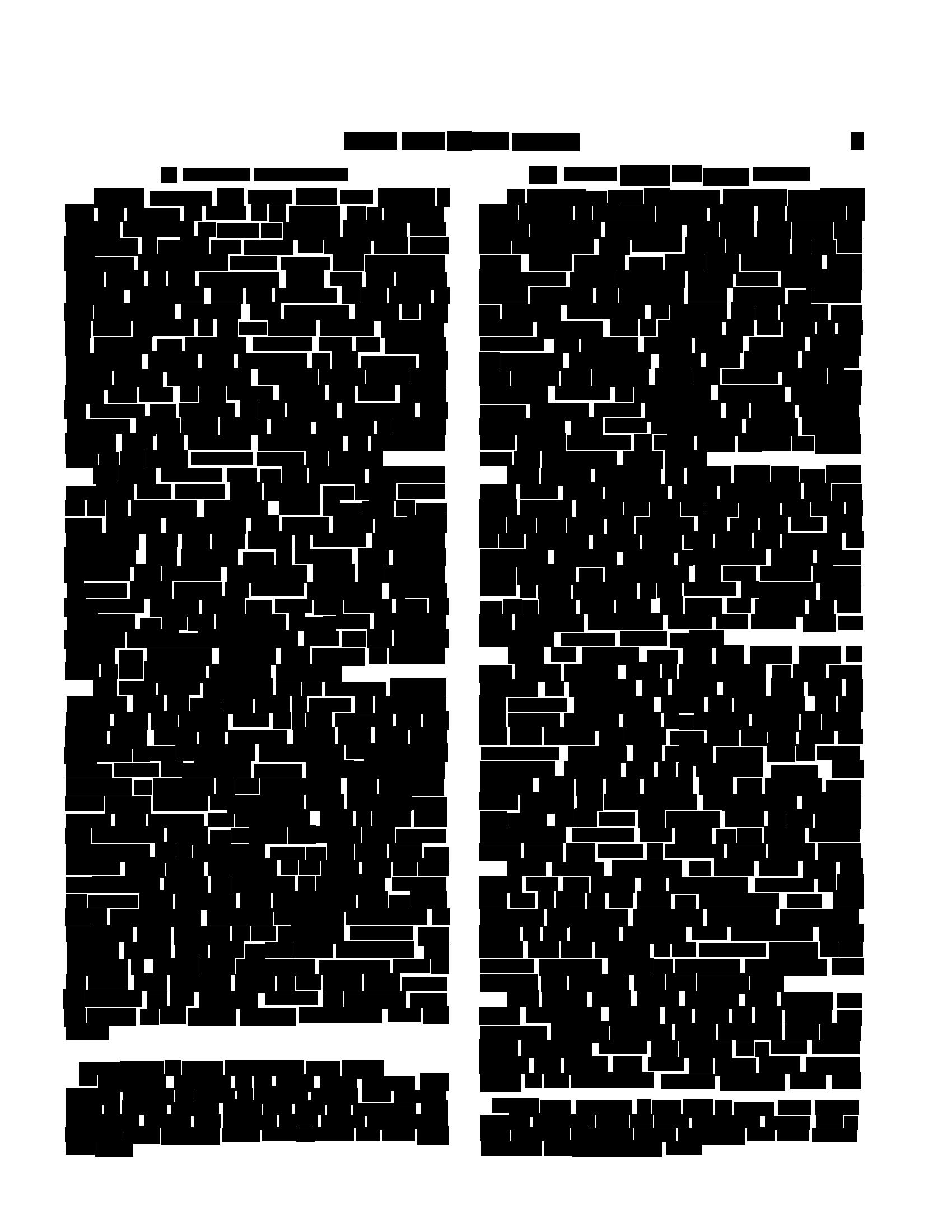}
        \hspace{1mm}
        \caption{DBNet based LF} 
        \label{fig:ex2}
     \end{subfigure}
     \hfill
     \begin{subfigure}[b]{0.15\textwidth}
        \centering
        \includegraphics[scale=0.03]{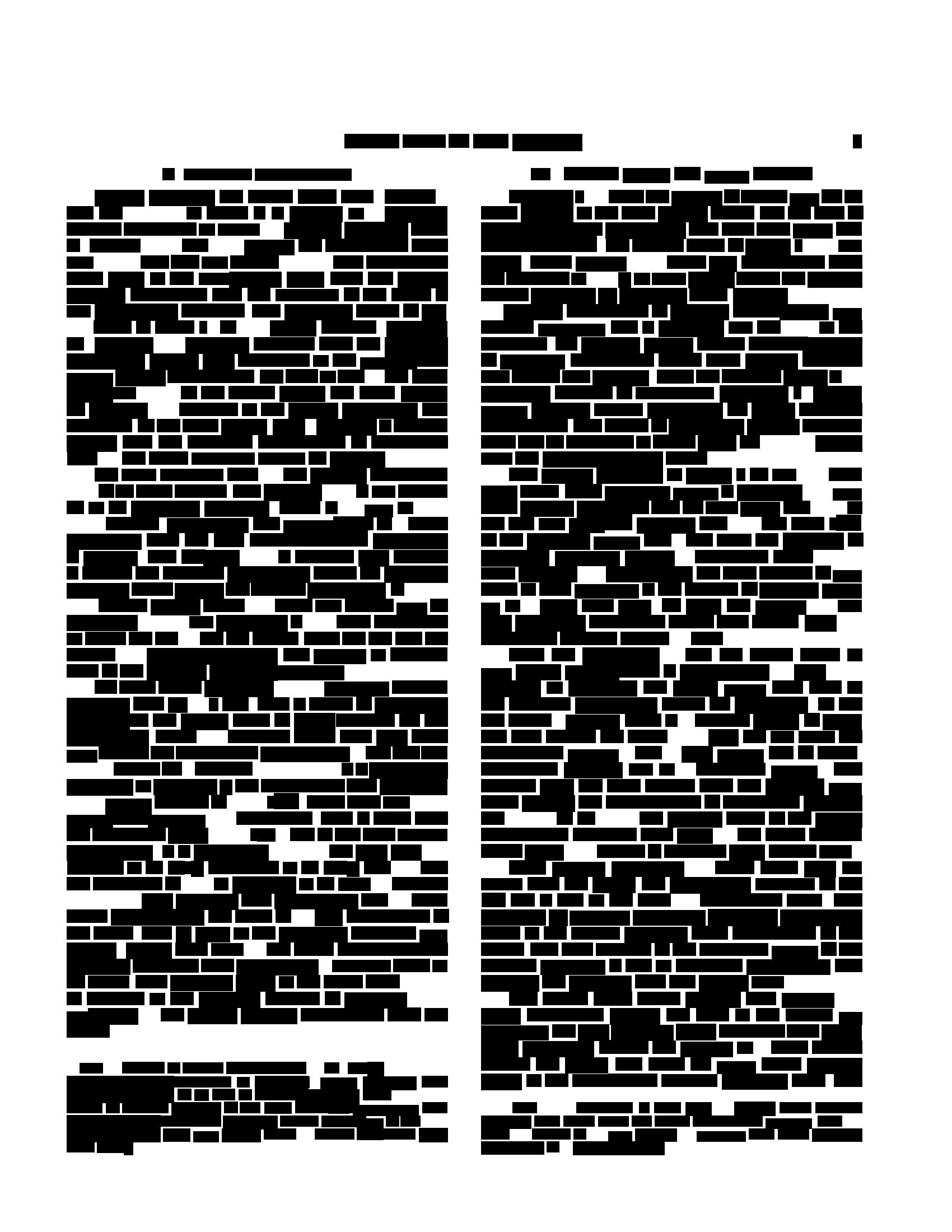}
        \hspace{1mm}
        \caption{Contour based LF} 
        \label{fig:ex3}
     \end{subfigure}
        \label{fig:example}
             \begin{subfigure}[b]{0.15\textwidth}
        \centering
        \includegraphics[scale=0.03]{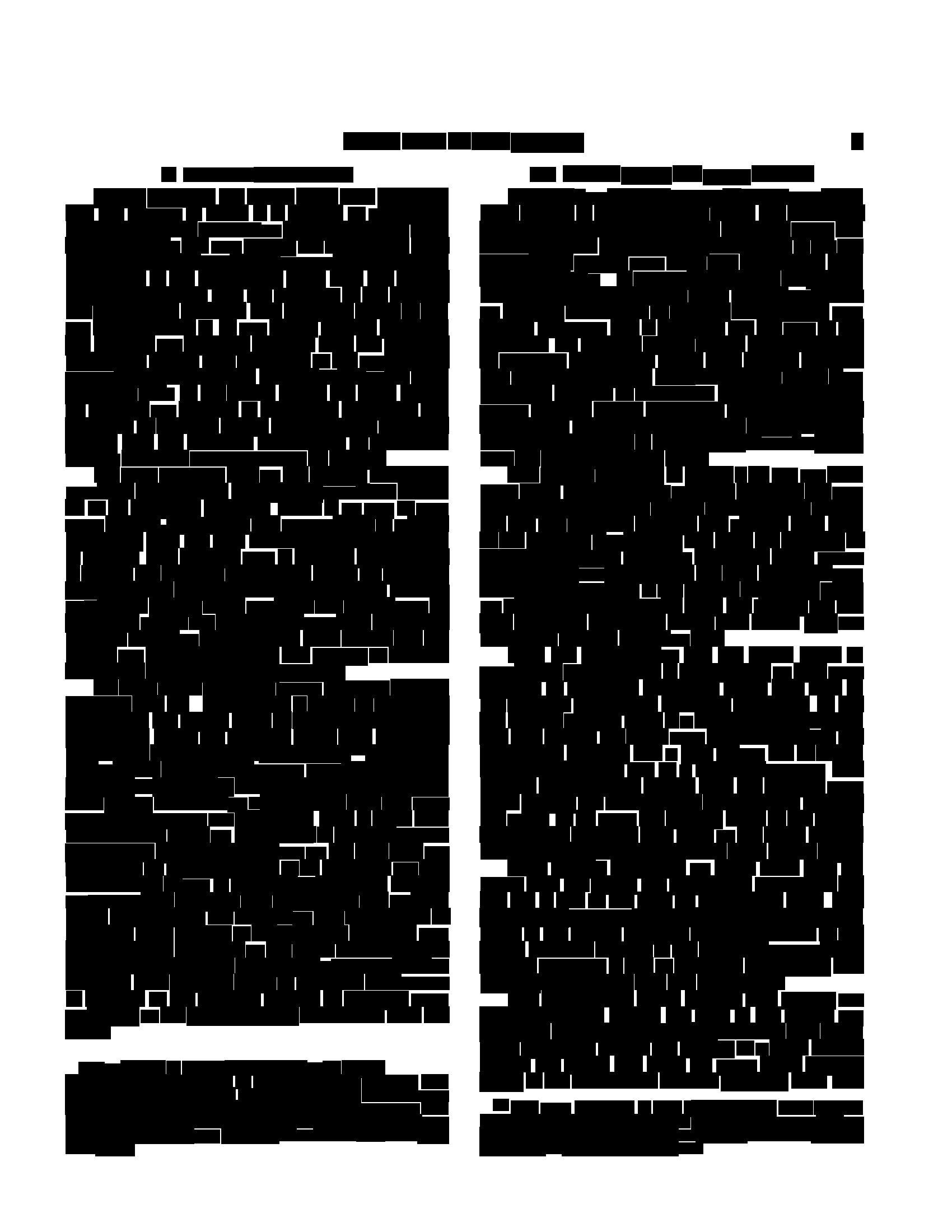}
        \hspace{1mm}
        \caption{\small{Canny Filter LF}} 
        \label{fig:ex4}
     \end{subfigure}
     \hfill
     \begin{subfigure}[b]{0.15\textwidth}
        \centering
        \includegraphics[scale=0.03]{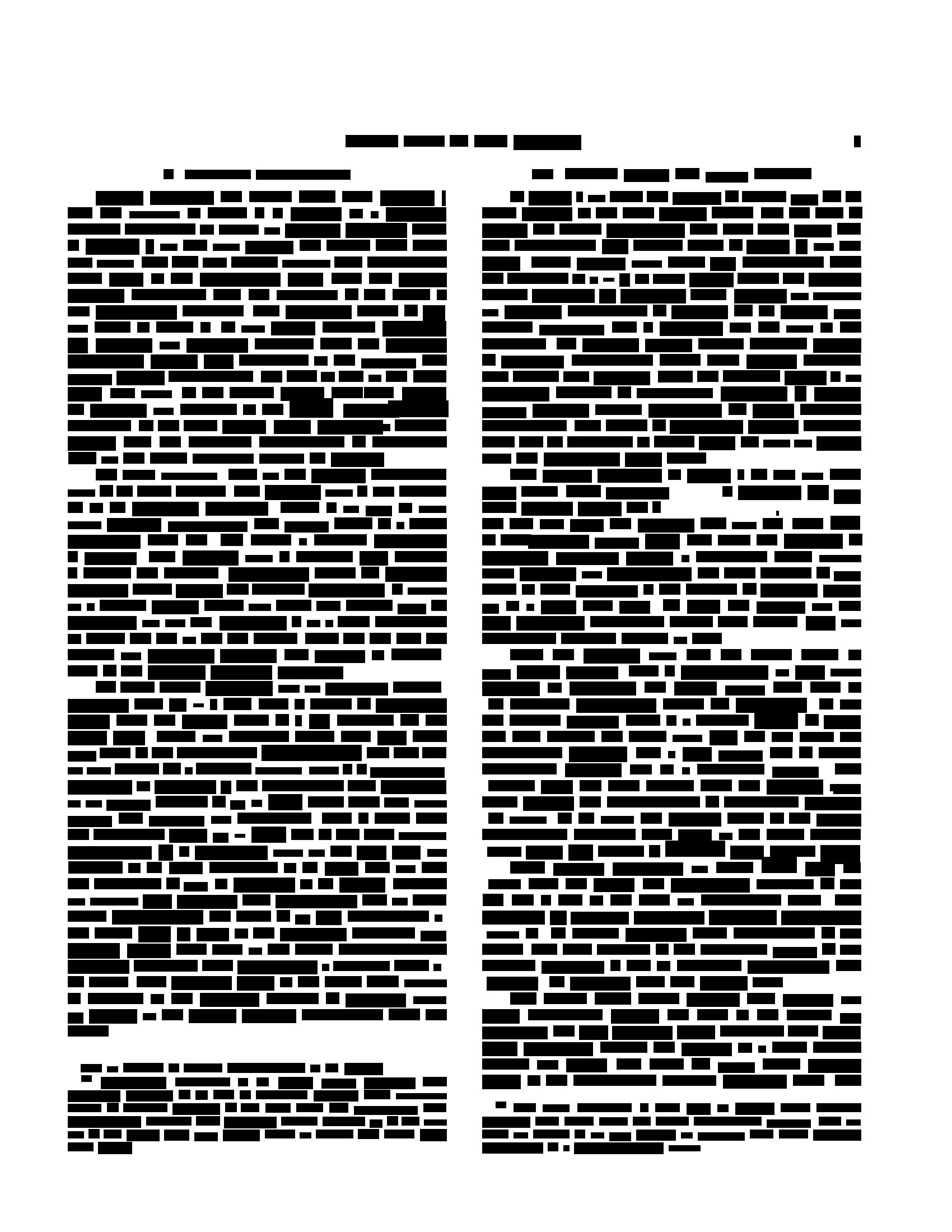}
        \hspace{1mm}
        \caption{Tesseract based LF} 
        \label{fig:ex5}
     \end{subfigure}
     \hfill
     \begin{subfigure}[b]{0.15\textwidth}
        \centering
        \includegraphics[scale=0.03]{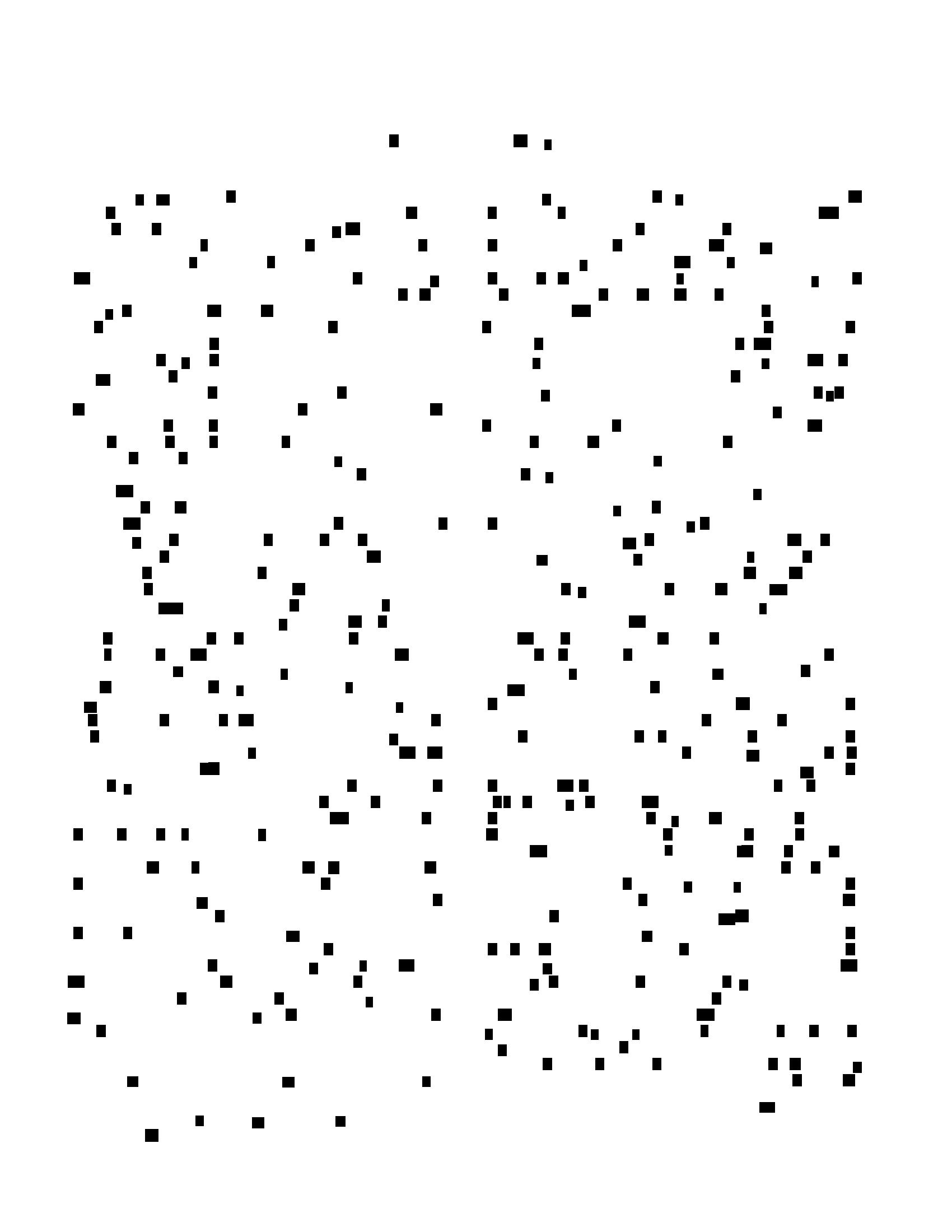}
        \hspace{1mm}
        \caption{Image Edges LF} 
        \label{fig:ex6}
     \end{subfigure}
        \caption{Binary Image Outputs for different Labeling Functions}
        \label{fig:example}
\end{figure*}

\begin{enumerate}
    \item \textbf{Labeling Functions from Pretrained DL Models} 
        \begin{enumerate}
            \item \textbf{DBNet~\cite{dbnet}} is pre-trained for English\footnote{\url{https://mindee.github.io/doctr/latest/modules/models.html\#doctr.models.detection.db\_resnet50}} and uses neural network segmentation-based method for text detection. Given a document image, DBNet internally generates a binarized pixel map which is then processed to generate the word-level bounding boxes. Based on these bounding boxes, this LF gives a binarized pixel map as its output as shown in Fig \ref{fig:ex2}
            
        \end{enumerate}
    \item \textbf{CV-based Labeling Functions}
        \begin{enumerate}
            \item \textbf{Contour-based LF}: Contour-based methods can determine word-level regions on a binarized pixel map by demarcating pixels having the same value. Given an input image (shown in Fig \ref{fig:ex1}), we binarize the image and find contours twice with an intermediate process of drawing contours\footnote{with thickness 4, which is a hyperparameter for this LF}. This is finally used to get the pixel-level binary map. The process to obtain bounding boxes is pictorially depicted in Figure \ref{contourlf}. The binarized output pixel map is shown in Fig \ref{fig:ex3}.

            \item \textbf{Canny Filter-based LF}: This function detects edges along the words by using the canny \cite{canny} filter over the input image. Later the output of detected edges is fed into the contour-based post-processing with certain edge thickness \footnote{used as a hyperparameter for this LF} as shown in Fig \ref{contourlf} to retrieve bounding boxes and obtain the final binarized output as shown in Figure~\ref{fig:ex4}.
            
            \item \textbf{Tesseract~\cite{smith2007overview}}: The image-to-data method provided by Tesseract\footnote{\url{https://github.com/tesseract-ocr/tessdoc}} can also determine word-level bounding boxes for text detection which is used to create a binarized output map. The output for the same is shown in Fig \ref{fig:ex5}

            \item \textbf{Image Edges based LF}: As shown in Figure~\ref{fig:ex6}, this function uses the Sobel filter~\cite{sobel} method to find the corresponding edges of the textual information present in a binarized input document image. The binary map is again fed to contour-based processing \footnote{with thickness 2, a hyperparameter for this LF} to obtain bounding boxes.
        \end{enumerate}
\end{enumerate}

\begin{figure}
     \centering
     \includegraphics[scale=0.30]{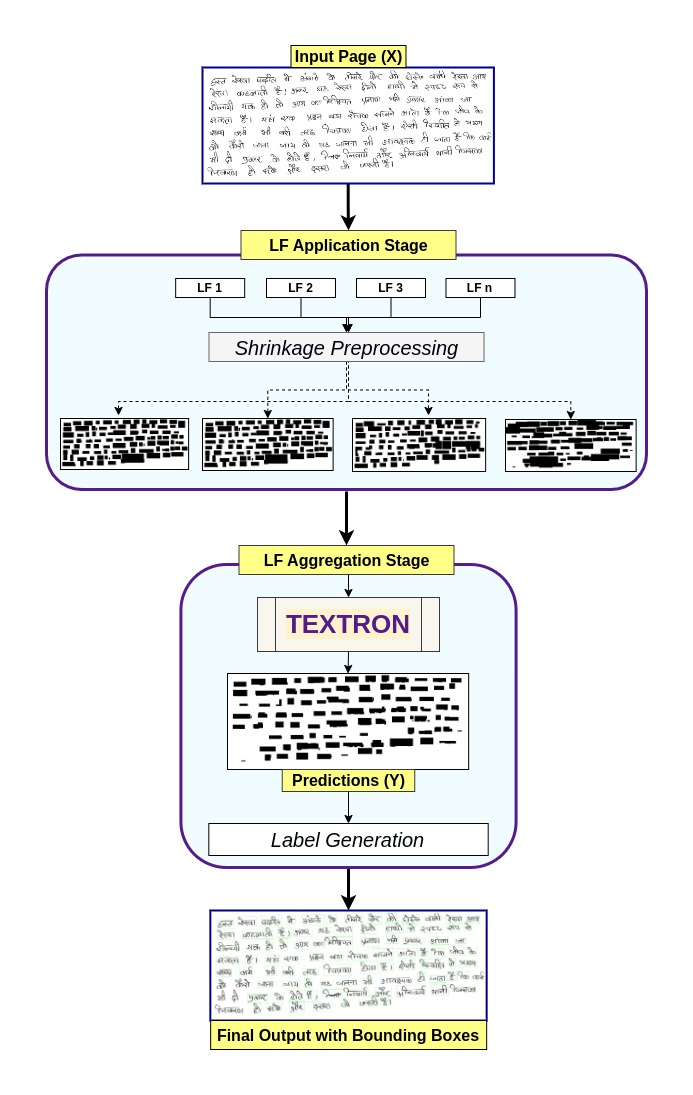}
      \caption{A step-by-step description of the \Textron{} workflow}
      \label{fig:workflow}
\end{figure}

Similar to the five LFs described above, we also define and use the corresponding five complementary LFs which are associated with the \nontext{} class. Thus, we have designed ten LFs out of which, the complementary LFs use the same algorithm proposed in the above descriptions. However, the only difference is that they help to label the concerned set of pixels with \nontext{} class. The reason to introduce the complementary LFs lies in the fact that currently, our method limits one labeling function to label entities for only a single class. The outputs for such complementary LFs will be the complement of what is shown in Figure \ref{fig:example} and will label the non-textual pixels of the input page image. The workflow of \Textron{ } is presented in Figure \ref{fig:workflow} which highlights the application of LFs followed by aggregation of LF outputs to produce \Textron{ } output, which is also a binarized pixel map (Y). Before proceeding to further details, we introduce two standard processes working in the periphery for the creation of word-level bounding boxes from the retrieved pixel-based \Textron{ } output.

\label{methodology}

% \begin{figure}
%      \centering
%      \begin{subfigure}[]{0.5\textwidth}
%         \centering
%         \includegraphics[scale=0.15]{images/lf_1.png}
%         \hspace{1mm}
%         \caption{Converting Input Image into a flattened array of pixels map} 
%         \label{fig:lf1}
%      \end{subfigure}
%      %\hfill
%      \begin{subfigure}[]{0.5\textwidth}
%         \centering
%         \includegraphics[scale=0.15]{images/lf_2.png}
%         \hspace{1mm}
%         \caption{Applying \Textron{} learning objective to the pixel map} 
%         \label{fig:lf2}
%      \end{subfigure}
%      %\hfill
%      \begin{subfigure}[]{0.5\textwidth}
%         \centering
%         \includegraphics[scale=0.15]{images/lf_3.png}
%         \hspace{1mm}
%         \caption{Label generation from the pixel map ${\mathbf Y}$} 
%         \label{fig:lf3}
%      \end{subfigure}
%      %\hfill
%         \caption{A step-by-step description of the \Textron{} workflow}
%         \label{fig:workflow}
% \end{figure}
% \label{methodology}

\subsubsection{Shrinkage Factor Preprocessing}
\begin{figure}
     \centering
     \begin{subfigure}[b]{0.15\textwidth}
        \centering
        \includegraphics[scale=0.12]{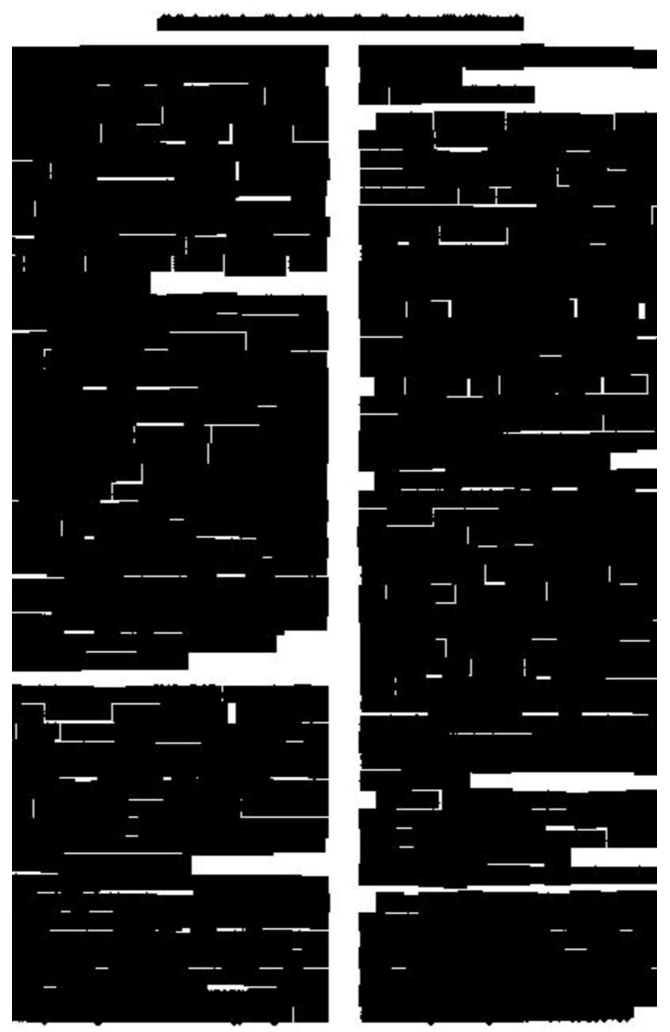}
        \hspace{1mm}
        \caption{Original Result without shrinking dimesnions} 
        \label{fig:shrink1}
     \end{subfigure}
     \hfill
     \begin{subfigure}[b]{0.15\textwidth}
        \centering
        \includegraphics[scale=0.12]{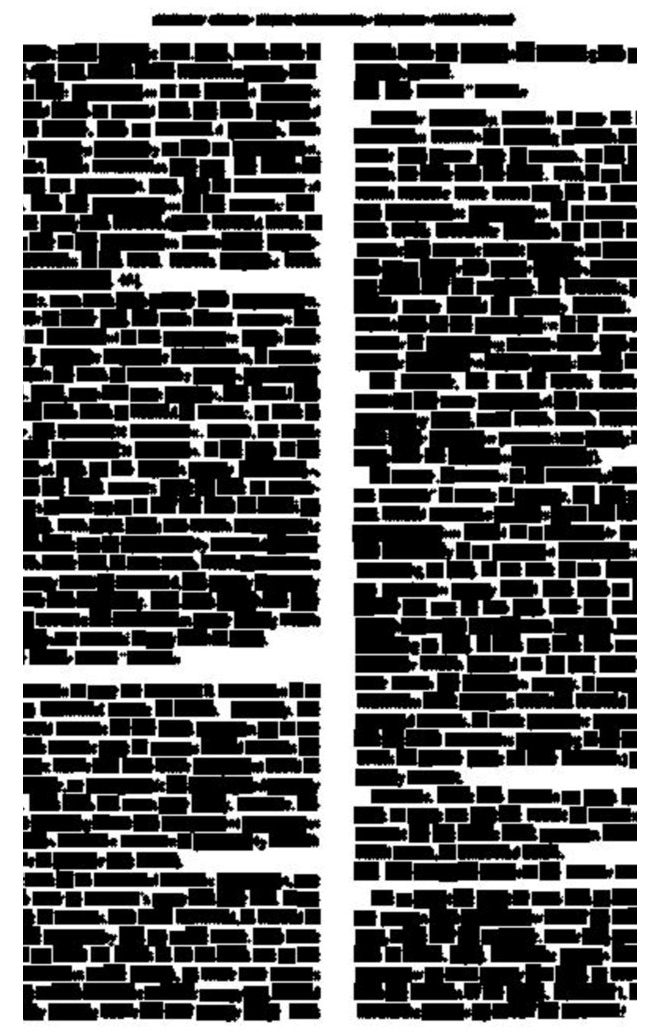}
        \hspace{1mm}
        \caption{With 20\% Shrinkage of both height and width} 
        \label{fig:shrink2}
     \end{subfigure}
     \hfill
     \begin{subfigure}[b]{0.15\textwidth}
        \centering
        \includegraphics[scale=0.12]{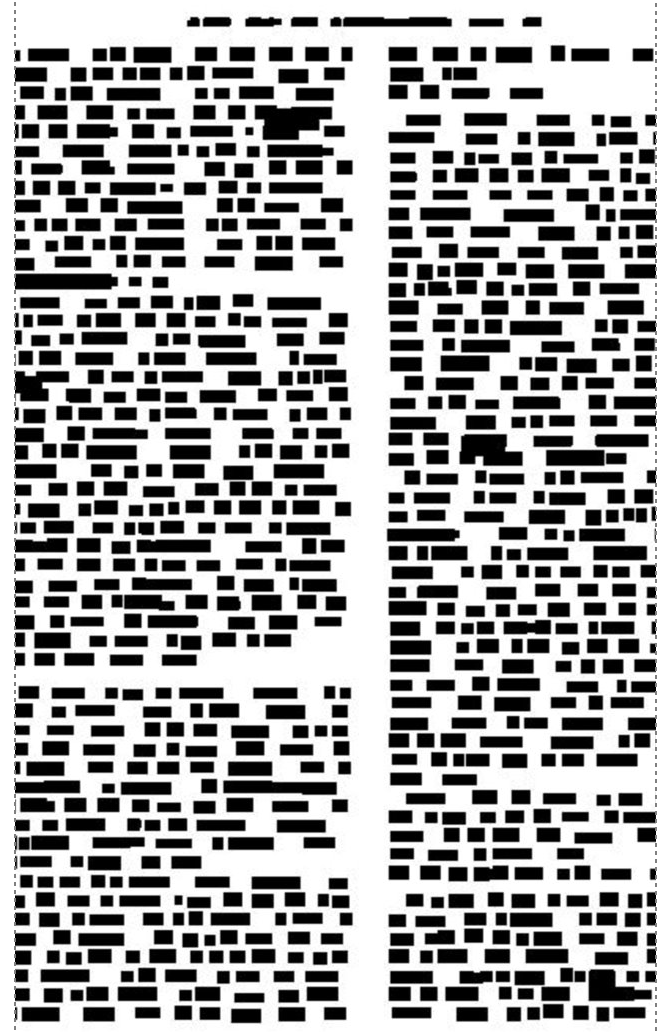}
        \hspace{1mm}
        \caption{With 40\% Shrinkage of both height and width} 
        \label{fig:shrink3}
     \end{subfigure}
        \caption{Demonstration of Shrinkage Factor Preprocessing}
        \label{fig:comparison}
\end{figure}

The final pixel-level prediction (Y) obtained in Figure~\ref{fig:workflow} is in the form of a binary map. Often, in such an output, we observe cases with overlapping bounding boxes as depicted in Figure \ref{fig:shrink1}. The word level demarcation gets lost in these overlapping bounding boxes. Since it is preferable to get the output in bounding boxes format, we need to tweak the LFs in such a way that we are able to retrieve the word level demarcation again through the output. To achieve this, we shrink the width and height of the bounding boxes provided by each LF by a shrinkage factor (set to be a hyperparameter) to provide a binary map containing disjoint non-overlapping bounding boxes. This encourages our \Textron{ } model to fit and learn on shrunk bounding boxes. This process, in turn, provides a similar output map as seen in Figures~\ref{fig:shrink2} and ~\ref{fig:shrink3} respectively. These figures show the effect on the overall results produced by shrinking heights and widths of bounding boxes with different values of shrinkage factors. The benefit of performing shrinkage becomes evident while recovering the original and independent word-level bounding boxes during the label generation step and this is explained in the following Section~\ref{label-generate}.

\subsubsection{The Label Generation Logic}
\label{label-generate}
\begin{figure}
     \centering
     \begin{subfigure}[b]{0.15\textwidth}
        \centering
        \includegraphics[scale=0.03]{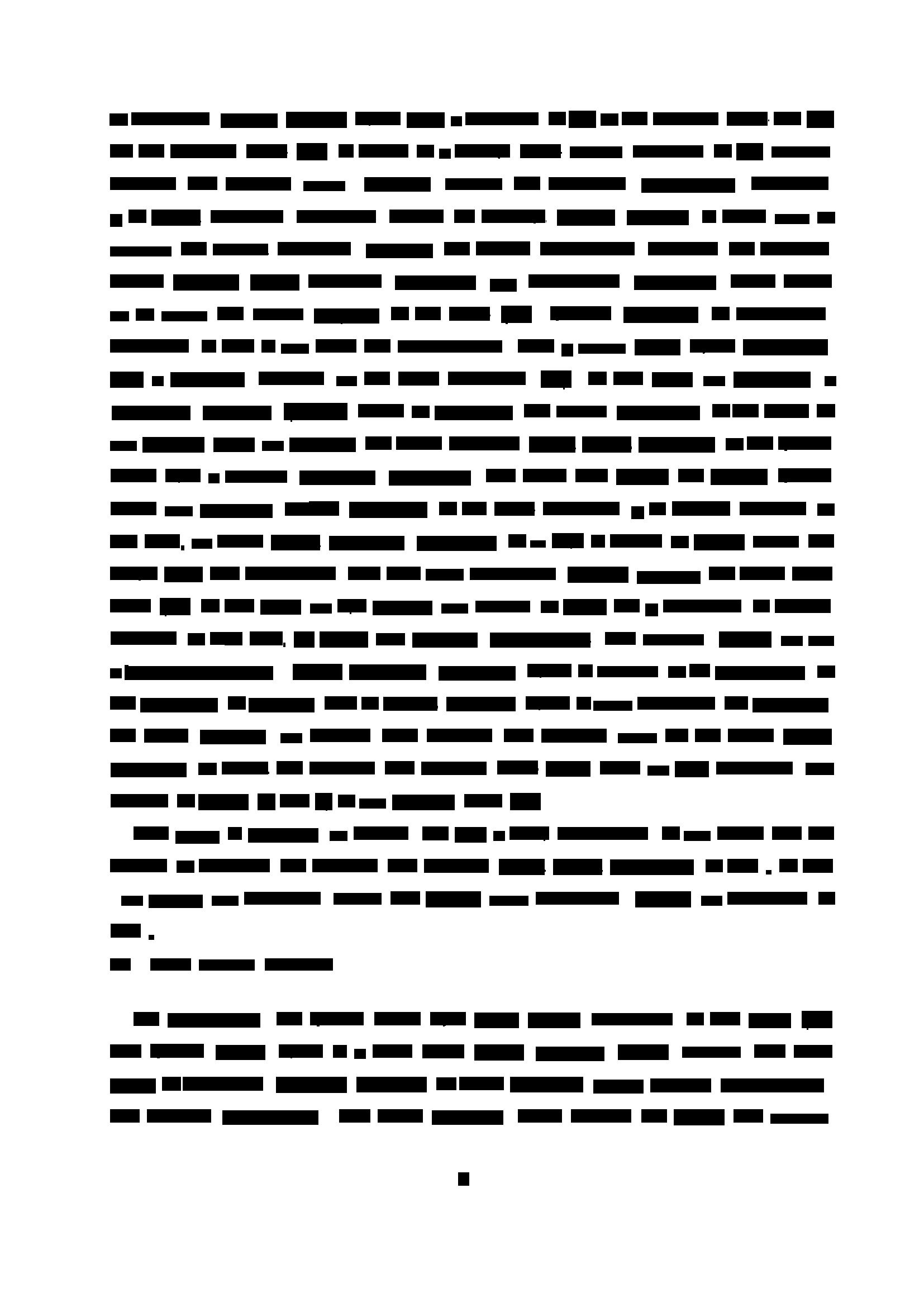}
        \caption{Textron Predictions} 
        \label{fig:postprocessing_1}
     \end{subfigure}
     \hfill
     \begin{subfigure}[b]{0.15\textwidth}
        \centering
        \includegraphics[scale=0.03]{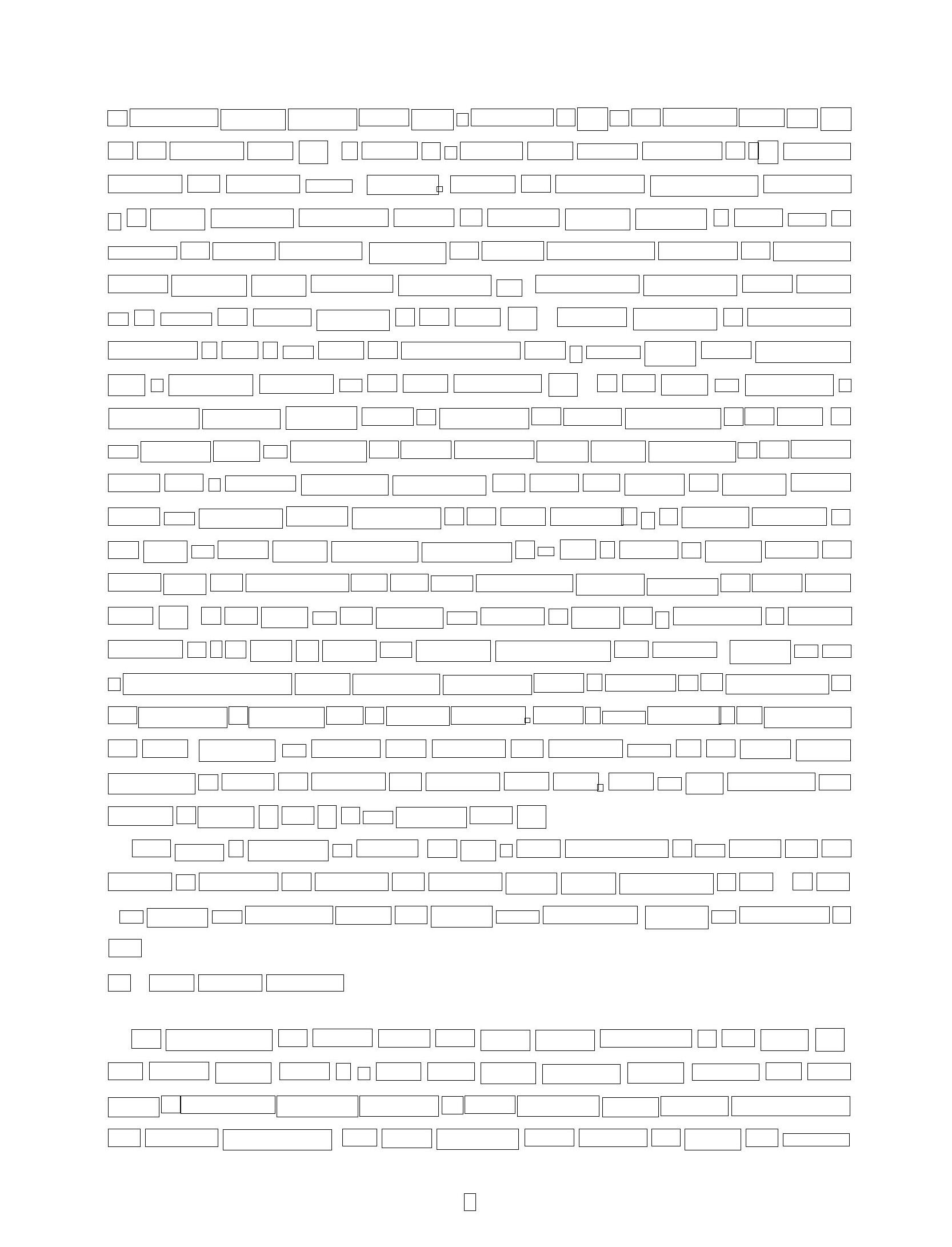}
        \caption{Contour-based processing} 
        \label{fig:postprocessing_2}
     \end{subfigure}
     \hfill
     \begin{subfigure}[b]{0.15\textwidth}
        \centering
        \includegraphics[scale=0.03]{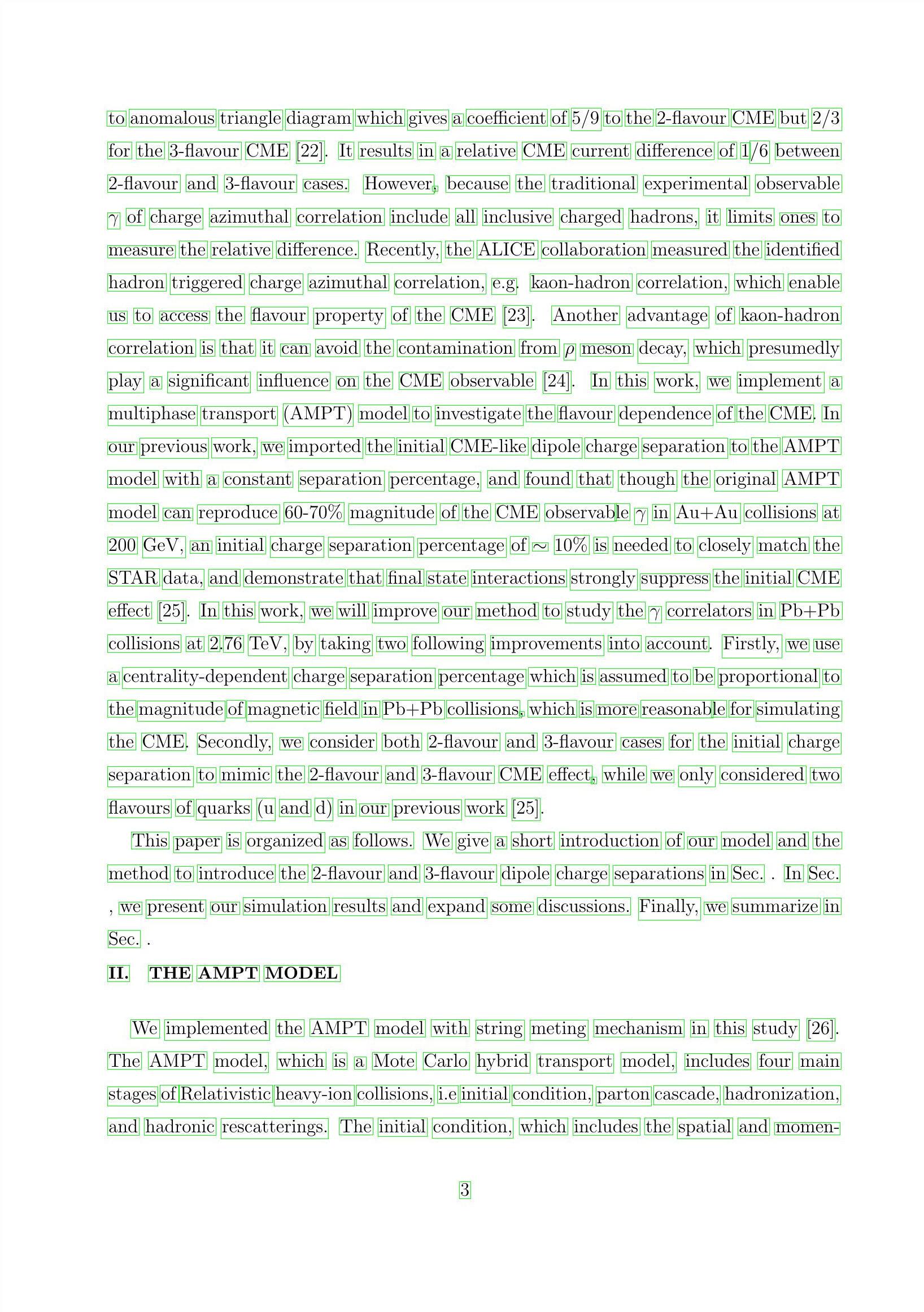}
        \caption{Final Predictions} 
        \label{fig:postprocessing_3}
     \end{subfigure}
        \caption{Contour-based post-processing step for label generation.}
        \label{fig:postprocessing}
\end{figure}

The generated binarized pixel map ($Y$ from Figure \ref{fig:workflow}) is subjected to a simple post-processing step to convert the pixel-level data into word-level bounding boxes. To achieve this, we use the  contour-based\footnote{Cv2 contours \url{https://docs.opencv.org/3.4/d4/d73/tutorial\_py\_contours\_begin.html}} 
demarcation method. After retrieving the corresponding bounding boxes, their widths and heights are recovered by enlarging them to the same extent by which they were shrunk in the LF application stage. In Figure~\ref{fig:postprocessing}, we present the stepwise workflow of label generation in order to retrieve the final predictions. Through this step, \Textron{ } is able to provide a sequence of bounding boxes as its output which can be further used for feedback provision or evaluation.

\subsection{Our Framework}

As mentioned in Section \ref{sec:intro}, our framework \Textron{ } tries to combine different text detection methods together using a data programming paradigm. The objective is to train a model that can learn the parameters associated with each LF so as to generate a binarized pixel map that aligns with the consensus of LF outputs. In our framework, the data programmer can guide the training process by providing quality guides to each LF~\cite{cage}, which are then incorporated into a generative model. This is important for our scenario because while designing LFs, a programmer is aware of the cases in which the LF (text detection method giving weak labels) might work efficiently or fail.
Let $\mathcal{X} =\{\mathbb{R}\}^{H \times W}$ where $\mathbb{X}$ represents binarized input image and  $\mathbb{R}$ represents the intensity or color value of the pixel, $H$ and $W$ represent the height and width of the image respectively. $\mathcal{Y}=\{c1,c2\}$ denotes the binary label space indicating whether or not a pixel represents a text region. $P(\mathcal{X},\mathcal{Y})$ denotes the joint distribution of pixels and labels. Our goal is to model the relation between a label \textit{y} with a pixel $x \in \mathcal{X}$. We have \textit{m=H*W} pixels for a given image instance. We have \textit{n} LFs $\lambda_1,\lambda_2,..,\lambda_n$ and each LF $\lambda_i$ is associated with a label $\textit{k}_i$ which is one of the labels \textit{c1} (for \textlabel{}) or \textit{c2} (for \nontext{}). $\lambda_i$ provides a discrete label $\tau_{ij}=\textit{k}_j$ when triggered and $\tau_{ij}=0$ when not triggered. Each LF corresponds to a CV or DL-based algorithm to detect textual and non-textual regions as long as it returns values in the above format. More details on the different LFs used are presented in Section \ref{lfs}. Given the different LFs, our goal is to learn to create consensus among the outputs of the LFs. Thus, the model imposes a joint distribution between the true label \textit{y} and the value $\tau_{ij}$ returned by each LF $\lambda_i$ on any pixel $x_j$. Since the graphical model (parameterized by $\theta$) is to be trained on unlabeled data, the model should ideally fit only on $\tau_{ij}$ without relying on the true label \textit{y}. Hence, the joint distribution is defined as,

\begin{equation}
    P_{\theta}\left(y, \tau_i\right)=\frac{1}{Z_\theta} \prod_{j=1}^n \psi_\theta\left(\tau_{i j}, y\right)
\end{equation}
where $\theta$ denotes the parameters used in defining the potentials $\psi_\theta$,
\begin{equation}
    \psi_\theta\left(\tau_{i j}, y\right)= \begin{cases}\exp \left(\theta_{j y}\right) & { if } \tau_{i j} \neq 0 \\ 1 & \text { otherwise }\end{cases}
\end{equation}
and the normalizer $Z_\theta$ will be
\begin{equation}
    Z_\theta = \sum_{y \in \mathcal{Y}} \prod_j\left(1+\exp \left(\theta_{j y}\right)\right)
\end{equation}
Given the above model, the training objective is defined as,
\begin{equation}
    \max _{\theta} L L(\theta \mid D)+R\left(\theta \mid\left\{q_j^t\right\}\right)
    \label{eq:lltrainobj}
\end{equation}
The first part maximizes the likelihood of the observed $\tau_i$ of the pixels in the training samples $D= x_1,., x_m$ after marginalizing out the true $y$. It can be expressed as: 
\begin{equation}
    \begin{aligned}
& L L(\theta \mid D)=\sum_{i=1}^m \log \sum_{y \in \mathcal{Y}} P_{\theta}\left(\tau_i, y\right) \\
& =\sum_{i=1}^m \log \sum_{y \in \mathcal{Y}} \prod_{j=1}^n \psi_j\left(\tau_{i j}, y\right)-m \log Z_\theta
\end{aligned}
\end{equation}

$R$ is a regularizer that guides the parameters with the programmer's expectation of the quality of each LF. The $q_j^t$ guide (a value between 0 and 1) is the user's belief or confidence in the fraction of the cases where $y$ and $t_j$ agree with $\tau_j \neq 0$. Given $q_j^t$ we seek to minimize the KL divergence between the user-provided $q_j^t$ and the model-calculated precision $P_\theta(y=k_j|\tau=k_j$) which turns out to be:
\begin{equation}
    \begin{array}{r}
R\left(\theta \mid\left\{q_j^t\right\}\right)=\sum_j q_j^t \log P_\theta\left(y=k_j \mid \tau_j=k_j\right) \\
+\left(1-q_j^t\right) \log \left(1-P_\theta\left(y=k_j \mid \tau_j=k_j\right)\right)
\end{array}
\end{equation}

In a similar manner, we proceed with the training by loading another unlabeled image, keeping the $\theta$ unchanged, and optimizing the same training objective mentioned in equation \eqref{eq:lltrainobj}. 
% Eventually, we keep loading a large number of unlabeled images and carry out detailed unsupervised analysis described in Section \ref{sec:analysis} to simultaneously tune the $\theta$ parameters for concerned LFs. 
It is during the provision of these unlabeled images that we can try to make up for the scarcity or absence of labeled data such as by providing unlabeled document page-level images having contents written in low-resource languages or handwritten Indian languages. After an acceptable number of image iterations, we save the model configuration which includes the LFs and their corresponding $\theta$ parameters. During the inference stage, we provide this model with quality guides and LF hyperparameters to perform text detection on the input image.     

For detailed motivation for the above equations, we refer the reader to the work of Chatterjee {\em et. al.}~\cite{cage}. Intuitively, given an image with $m$  pixels and $n$ LFs (and its corresponding $q_j^t$ quality guides), we train a graphical model by maximizing the log-likelihood training objective specified in equation \eqref{eq:lltrainobj}. 

\section{Experiments}
\label{sec:experiments}
To demonstrate the efficacy of the proposed framework, we ran extensive experiments. We begin by describing the datasets used and then present our experimental setup.

\subsection{Datasets}
\label{dataset}
To compare the results in both the supervised and unsupervised settings, we selected two types of datasets. Publicly available datasets for the English language and text detection datasets for a few Indian languages for which there are no pre-trained models available, to the best of our knowledge. 

\subsubsection{English Datasets} \label{dataset-Docbank} 
Docbank\cite{docbank} is a publicly available dataset on which we benchmark our experimental results. The dataset contains around 500,000 page images each of which has corresponding annotations in the form of bounding boxes. Each bounding box is associated with a class. For our purpose, out of the 12 Docbank classes, we excluded the figure class (\ie, the 'figure' class will be considered as any other non-textual region) and considered the remaining 11 classes as the target, namely Abstract, Author, Caption, Equation, Footer, List, Paragraph, Reference, Section, Table, and Title. All regions belonging to the aforementioned classes are considered to be \textlabel{} regions. Out of the 500,000 images, we used around 22,000 images for training, validation, and unsupervised assessment of LFs. We also use a sample test set of 100 images having 55,223 bounding boxes representing text class on which we report our results. We also use 199 images from Funsd \cite{funsd} and 200 random images from the CTDAR dataset \cite{icdar2019} for training and validation purposes.

\subsubsection{Indian Language Text Datasets} 
\label{dataset-indian}
We have created a benchmark annotated dataset of printed document images in three Indian languages which we use as a test set. This includes 200 pages of Malayalam (48,358 words), 225 pages of Tamil (48,239 words), and 323 pages of Gujarati (82,475 words) documents. The Gujarati document pages also contain a considerable amount of Sanskrit text. We have collected the required pages from textbooks \cite{maltextbook} and other books \cite{geeta}. Word-level bounding boxes are marked for every page-level image by a group of qualified annotators. Apart from that, we also split and used around 300 unlabeled Sanskrit images from a set of books \cite{geeta} for training and validation respectively. The PHD Indic 11 dataset\cite{phdindic} is composed of handwritten document images of 11 official Indian language scripts. We have used 220 page-level images of handwritten Devanagari documents as our test set for which bounding boxes are marked by a group of verified annotators. We also use unlabeled document images of Telugu text (85 pages) and Kannada text (46 pages) from the PHD Indic dataset \cite{phdindic} for training, validation, and unsupervised analysis.

\subsection{Unsupervised Quantitative Assessment of LFs}
\label{sec:analysis}
\begin{figure}
    \centering
    \includegraphics[width=0.5\textwidth]{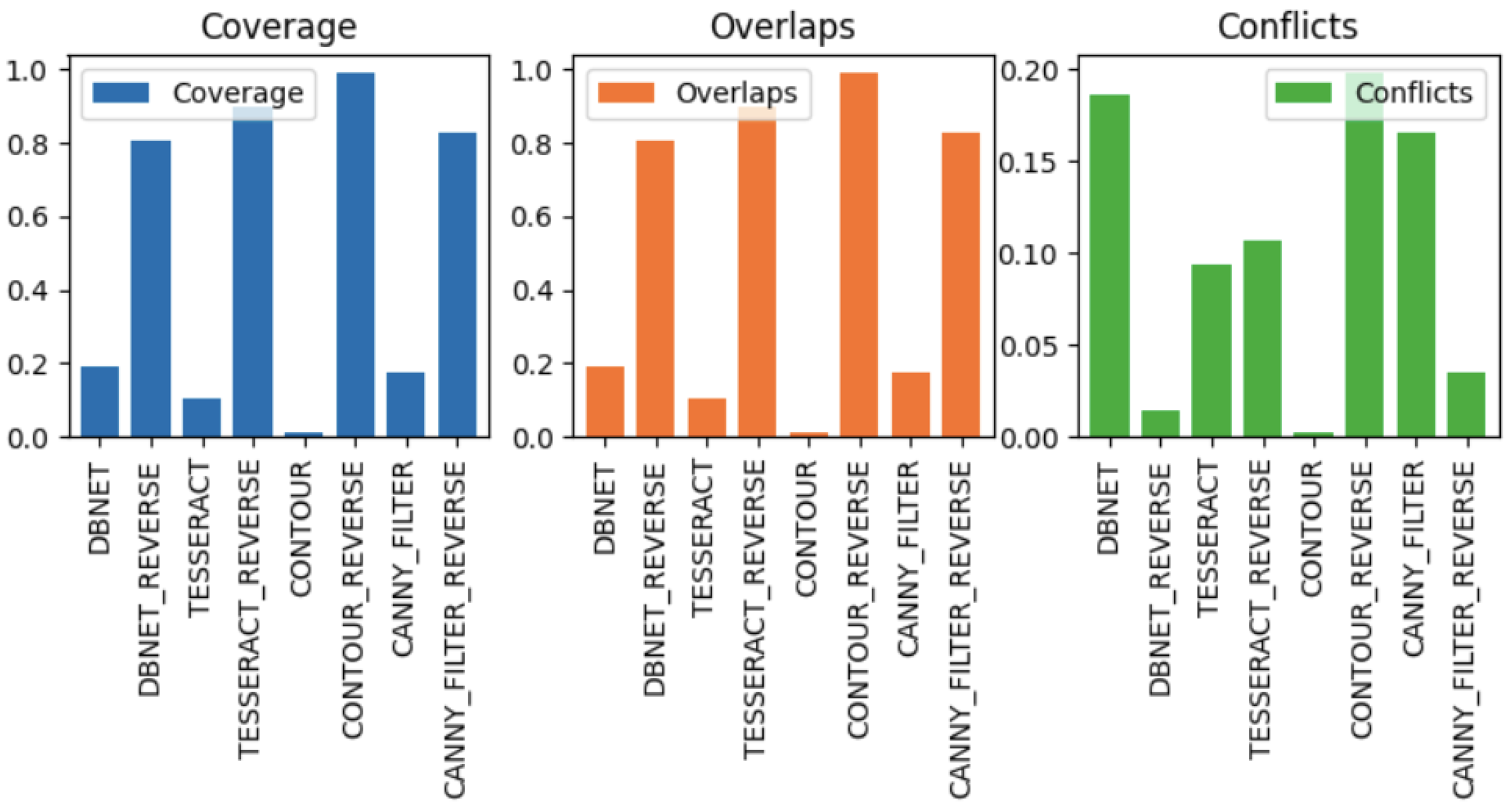}
    \caption{Unsupervised quantitative assessment on the eight LFs included in the experimentation displaying coverage, overlaps, and conflicts for each LF} 
    \label{fig:lfanalysis}
\end{figure}

While choosing the best-performing \Textron{ } LF set, we performed extensive experiments using different combinations of CV-based LFs as well as the LFs derived from pre-trained DL models. We analyzed the unsupervised performance measures of the ten LFs designed as mentioned in Section \ref{lfs}. These measures include coverage, overlaps, and conflicts for each LF. Recall from Section \ref{methodology} that each LF can either get triggered or abstain from labeling a pixel. The fraction of pixels (with respect to the complete set of pixels of the image) an LF can label is what we refer to as the {\em coverage} of each LF. Similarly, the {\em overlap} of an LF is defined as the fraction of pixels out of all covered (labeled) pixels assigned the same label by at least one more LF. As opposed to that, we define the {\em conflict} of an LF as the fraction of pixels with respect to the overlapped pixels that do not match the label assigned by any other LFs. Ideally, a good LF will have significant coverage, high overlap, and fewer conflicts. While experimenting we saw that the Image Edge-Based LF had an extremely low coverage for \textlabel{} pixels. Also, the corresponding complementary LF had high conflicts. So we dropped this LF and its chttps://github.com/IITB-LEAP-OCR/TEXTRONorresponding complementary LF from the best-performing set and analyzed further for the remaining eight LFs. These eight LFs included four fundamental LFs namely pre-trained DBNet, Tesseract-based LF, Contour-based LF, and Canny Filter-based LF along with their complementary LFs. Figure \ref{fig:lfanalysis} presents the three unsupervised performance measures for these LFs on a random image of the Docbank dataset. The fundamental LFs that labeled the text pixels had less (yet significant) coverage which was a favorable scenario as the amount of \textlabel{} pixels in any generic document page is much less as compared to the \nontext{} pixels. Similarly, all our chosen complementary LFs have a high percentage of coverage and overlap, both of which are above 80\% for almost every image. We have performed a similar kind of analysis with various unlabeled images with handwritten and printed Indian language text training sets as well. All kinds of analysis and visual inspection of outputs (presented in the supplementary) have pointed in favor of usage of $\Textron_{8LF}$. With this newly defined $\Textron_{8LF}$ configuration, we train our graphical model (parameterized by $\theta$) for the training objective described in equation \eqref{eq:lltrainobj}.

\subsection{Training and Validation}
\begin{figure}
  \begin{subfigure}[b]{0.5\linewidth}
    \centering
    \includegraphics[width=\linewidth]{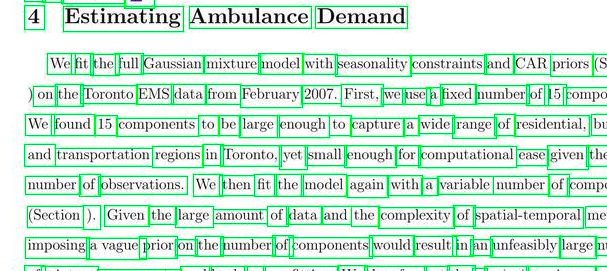} 
    \caption{Docbank Image} 
  \end{subfigure}%%
  % \begin{subfigure}[b]{0.5\linewidth}
  %   \centering
  %   \includegraphics[width=\linewidth]{images/train2.png} 
  %   \caption{Handwritten Telugu} 
  % \end{subfigure} 
  \begin{subfigure}[b]{0.5\linewidth}
    \centering
    \includegraphics[width=\linewidth]{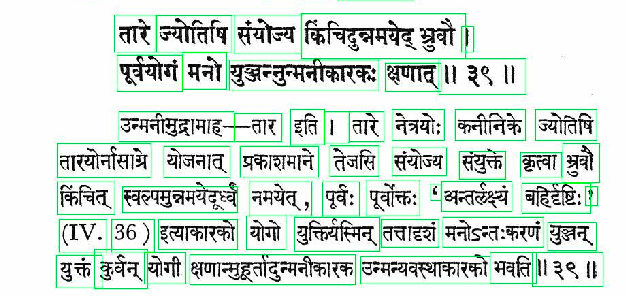} 
    \caption{Printed Sanskrit} 
  \end{subfigure}%% 
  % \begin{subfigure}[b]{0.5\linewidth}
  %   \centering
  %   \includegraphics[width=\linewidth]{images/train4.png} 
  %   \caption{CTDAR Image} 
  %   % \vspace{4ex}
  % \end{subfigure} 
    \caption{Predictions during $\Textron_{8LF}$ Learning Phase} 
    \label{fig:predictions}
\end{figure}

After unsupervised experimentation with various combinations of LFs on the training sets described in Section \ref{lfs}, we trained the $\Textron_{8LF}$ configuration. As mentioned in Section \ref{sec:methodology}, we try to optimize the training objective for 50 epochs by keeping a constant learning rate of 0.01 for each image in the train set. We further experimented on our various validation sets described in Section \ref{dataset}. This included tweaking the LF quality guides and their hyperparameters for an already trained $\Textron_{8LF}$ model. Hyperparameter tuning was performed on curated validation sets of 125 images (English), 85 images (Printed Sanskrit), and 90 images (Kannada and Telugu handwritten) to determine the best thickness and shrinkage parameters. Figure \ref{fig:predictions} shows our final word-level bounding boxes output on random images from the validation set produced by $\Textron_{8LF}$. Once the model was able to perform well through hyperparameter tuning for a sufficiently large number of images, we eventually used this saved $\Textron_{8LF}$ model and tuned hyperparameters for inference on unseen test sets.

\section{Results and Discussions}
\label{sec:results}
In this section, we present the performance of our $\Textron_{8LF}$ model using various combinations of quality guides and hyperparameters on the various test sets described in Section \ref{dataset}. 

\subsection{Baselines}
 We use the current state-of-the-art DBNet \cite{dbnet} model for text detection as our baseline. Additionally, we also use Majority Based Voting (MBV) as a baseline to compare our performance. The MBV baseline performs pixel-based voting through exactly the same combination of LFs (taking into account a pair of both fundamental LF and its corresponding complementary LF) used by $\Textron_{8LF}$. Each LF will cast a vote for a pixel either as \textlabel{} or \nontext{}. Finally, if the number of \textlabel{} votes for a pixel exceeds the \nontext{} votes, then the pixel is assigned \textlabel{} class. In this manner, MBV produces a binarized image that undergoes contour-based post-processing as described in Section \ref{label-generate} to obtain the bounding boxes. We evaluate $\Textron_{8LF}$ and the baselines using an IOU-based approach to determine the overall precision, recall, and F1 scores.

 \begin{figure*}
     \centering
     \begin{subfigure}[b]{0.24\textwidth}
        \centering
        \includegraphics[scale=0.3]{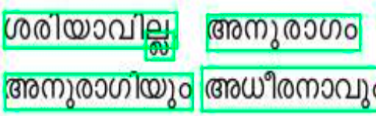}
        \hspace{1mm}
        \caption{DBNet Predictions for Malayalam} 
        \label{fig:err-mal}
     \end{subfigure}
     \hfill
     \begin{subfigure}[b]{0.24\textwidth}
        \centering
        \includegraphics[scale=0.28]{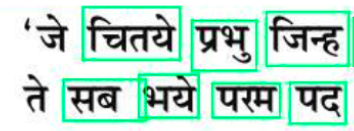}
        \hspace{1mm}
        \caption{DBNet Predictions for Sanskrit} 
        \label{fig:err-sans}
     \end{subfigure}
     \hfill
     \begin{subfigure}[b]{0.24\textwidth}
        \centering
        \includegraphics[scale=0.27]{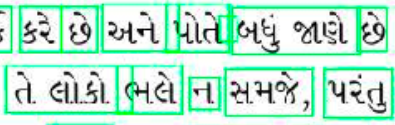}
        \hspace{1mm}
        \caption{DBNet Predictions for Gujarati} 
        \label{fig:err-guj}
     \end{subfigure}
    \begin{subfigure}[b]{0.24\textwidth}
        \centering
        \includegraphics[scale=0.16]{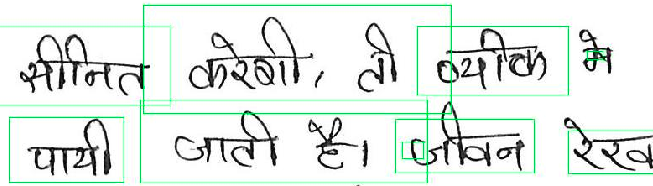}
        \hspace{1mm}
        \caption{DBNet Predictions for Hindi} 
        \label{fig:err-dev}
     \end{subfigure}
        
    \begin{subfigure}[b]{0.24\textwidth}
        \centering
        \includegraphics[scale=0.3]{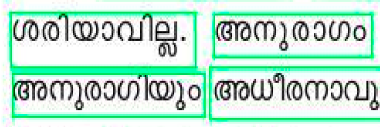}
        \hspace{1mm}
        \caption{Our Predictions for Malayalam} 
        \label{fig:res-mal}
     \end{subfigure}
     \hfill
     \begin{subfigure}[b]{0.24\textwidth}
        \centering
        \includegraphics[scale=0.28]{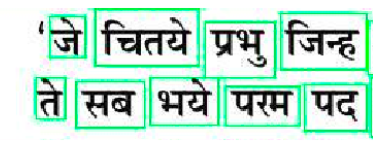}
        \hspace{1mm}
        \caption{Our Predictions for Sanskrit} 
        \label{fig:res-sans}
     \end{subfigure}
     \hfill
     \begin{subfigure}[b]{0.24\textwidth}
        \centering
        \includegraphics[scale=0.27]{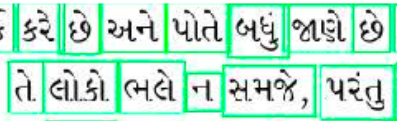}
        \hspace{1mm}
        \caption{Our Predictions for Gujarati} 
        \label{fig:res-guj}
     \end{subfigure}
          \begin{subfigure}[b]{0.24\textwidth}
        \centering
        \includegraphics[scale=0.24]{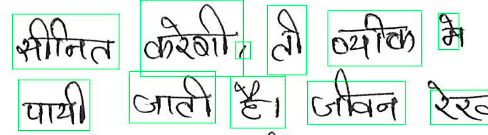}
        \hspace{1mm}
        \caption{Our Predictions for Hindi} 
        \label{fig:res-dev}
     \end{subfigure}
        \caption{Comparative Overview of DBNet results and our \Textron{ } outputs for different kinds of test images}
        \label{fig:error-resolution}
\end{figure*}

\subsection{Hyperparameters for Inference}
In this section, we mention the best-performing set of hyperparameters for $\Textron_{8LF}$ that have been tuned on validation sets. The fundamental four LFs used namely DBNet-based LF, Contour-based LF, Tesseract-based LF, and Canny Filter-based LF have quality guides set to 0.9, 0.85, 0.75, and 0.85 respectively. Quality guides of the complementary LFs are set to be 0.95 each. For this configuration, we apply 10\% width shrinkage as the preprocessing during the LF application stage. Additionally, the hyperparameters of \textit{height shrinkage},  \textit{contour thickness}, and \textit{edge thickness} were determined by tuning on different Indian language validation sets (described in Section \ref{dataset}). They are mentioned in Table \ref{tab:hyperparam}. Using these $\Textron_{8LF}$ configurations, we infer and report the following results on unseen test sets.

\begin{table}[]
    \centering
    \begin{tabular}{ | c | c | c | c | } 
  \hline
  \textbf{Dataset} & \textbf{\makecell{Height \\ Shrunk By}} & \textbf{\makecell{Contour \\ Thickness}} & \textbf{\makecell{Edge \\ Thickness}} \\ 
  \hline
   Docbank & 20\% & 4 & 2 \\ 
  \hline
   Malayalam & 20\% & 5 & 4 \\ 
  \hline
    Tamil & 20\% & 4 & 3 \\ 
  \hline
  Gujarati & 20\% & 4 & 3 \\ 
  \hline
  Devanagari & 30\% & 5 & 5 \\ 
  \hline
\end{tabular}
\caption{Hyperparameters of $\Textron_{8LF}$ for different datasets}
    \label{tab:hyperparam}
\end{table}

%\subsection{Performance on Docbank Dataset}

% Our approach performs better in terms of Recall and F1-Score and is at par with the best precision of the DBNet baseline. 

% Additionally, in Table \ref{tab:2}, we present a comparison of the classwise F1 scores of the word-level bounding boxes detected by the DBNet and MBV baseline against  $\Textron_{8LF}$. We have carried out this evaluation with an IOU threshold of 0.5. Our model not only performs better for half of
% the classes but also shows an overall improvement compared to the baseline performances. However, $\Textron_{8LF}$ performs poorly on the title class as titles generally have a different and larger font; CV-based LFs fail to unify title words in one bounding box. We can tweak the quality guides or include more relevant LFs in the paradigm to work on such limitations. Besides, there is also a class imbalance observed in the
% test set since classes such as date, author and title make up less than 1 percent of all bounding boxes, owing to which performance drop for those classes seems significant. 

\begin{table}[]
    \centering
    \begin{tabular}{ | c | c | c | c | } 
  \hline
  \textbf{Approach} & \textbf{P} & \textbf{R} & \textbf{F}\\ 
  \hline
   DBNet &  \textbf{90.49} & 80.00 & 84.92
 \\ 
  \hline
   Tesseract &  40.49 & 74.03 & 52.35
  \\ 
  \hline
    MBV &  89.53 & 80.02 & 84.51 \\ 
  \hline
  $\Textron_{8LF}$ &  90.17 & \textbf{80.75} & \textbf{85.21}
  \\ 
  \hline
\end{tabular}
\caption{Results with IOU 0.5 on Docbank 100 Test Samples}
    \label{tab:1}
\end{table}

\begin{table*}[]
\centering
    \begin{tabular}{ || c || c | c | c || c | c | c || c | c | c || c | c | c || } 
  \hline
  Language & \multicolumn{3}{|c||}{Printed Malayalam} & \multicolumn{3}{|c||}{Printed Tamil} & \multicolumn{3}{|c||}{Printed Gujarati} & \multicolumn{3}{|c||}{Handwritten Devanagari} \\
  \hline
  \textbf{Approach} & \textbf{P} & \textbf{R} & \textbf{F} & \textbf{P} & \textbf{R} & \textbf{F}  & \textbf{P} & \textbf{R} & \textbf{F} & \textbf{P} & \textbf{R} & \textbf{F} \\ 
  \hline
   DBNet &  97.97 & 99.38 & 98.67 & 98.90 & 99.59 & 99.24 & 98.40 & 96.84 & 97.62  &  \textbf{72.84} & 62.33 & 67.17\\ 
  \hline
  Tesseract &  81.39 & 90.96 & 85.91 &  95.23 & 95.34 & 95.28  &  86.72 & 84.06 & 85.37  & 60.29 & 65.89 & 62.97\\
  \hline
  MBV &   \textbf{99.71} & 99.35 & \textbf{99.53}  &   99.29 & 99.25 & 99.27  &   95.39 & \textbf{98.38} & 96.86 & 57.05 & 68.76 & 62.36 \\
  \hline
  $\Textron_{8LF}$ & 99.03 & \textbf{99.44} & 99.23 & \textbf{99.32} & \textbf{99.64} & \textbf{99.48} & \textbf{98.57} & 97.65 & \textbf{98.11} & 69.24 & \textbf{74.51} & \textbf{71.78} \\
  \hline
\end{tabular}
    \caption{Results with IOU 0.5 on Indian Language Text Detection Test Sets}
     \label{tab:indian-perf}
\end{table*}

% \begin{center}
% \setlength{\tabcolsep}{0.2em}
% \begin{table}
% \def\arraystretch{1.2}%
% \centering
% \begin{tabular}{| c | c | c | c | c |}
% \hline
% {\textbf{ Class }} & \textbf{Samples} &  {\textbf{DBNet}} & {\textbf{MBV}} & {\small{\textbf{ $\Textron_{8LF}$}}} \\
% \hline
% Date          & 9     & 100.00 & 100.00& 100.00\\
% Author        & 45    &  98.88 & 88.17 & 96.70 \\
% Title         & 71    &  91.18 & 54.19 & 75.00 \\
% Section       & 435   &  92.63 & 81.73 & 97.45 \\
% List          & 478   &  87.29 & 89.03 & 88.40 \\
% Abstract      & 740   &  91.30 & 92.51 & 92.68 \\
% Footer        & 870   &  91.35 & 86.01 & 96.00 \\
% Caption       & 1317  &  87.98 & 89.67 & 88.31 \\
% Table         & 2669  &  58.24 & 50.51 & 50.59 \\
% Equation      & 4190  &  17.66 & 32.54 & 22.08 \\
% Reference     & 5571  &  94.38 & 94.07 & 94.18 \\
% Paragraph     & 38828 &  89.39 & 88.76 & 90.12 \\ \hline
% Overall       & 55223 &  84.92 & 84.51 & 85.21 \\ \hline
% \end{tabular}
% \caption{Docbank Classwise F1 Scores (in \%) for IOU 0.5}
% \label{tab:2}
% \end{table}
% \end{center}

\subsection{Performance on Different Datasets}
\begin{figure}
    \centering
    \includegraphics[width=0.35\textwidth]{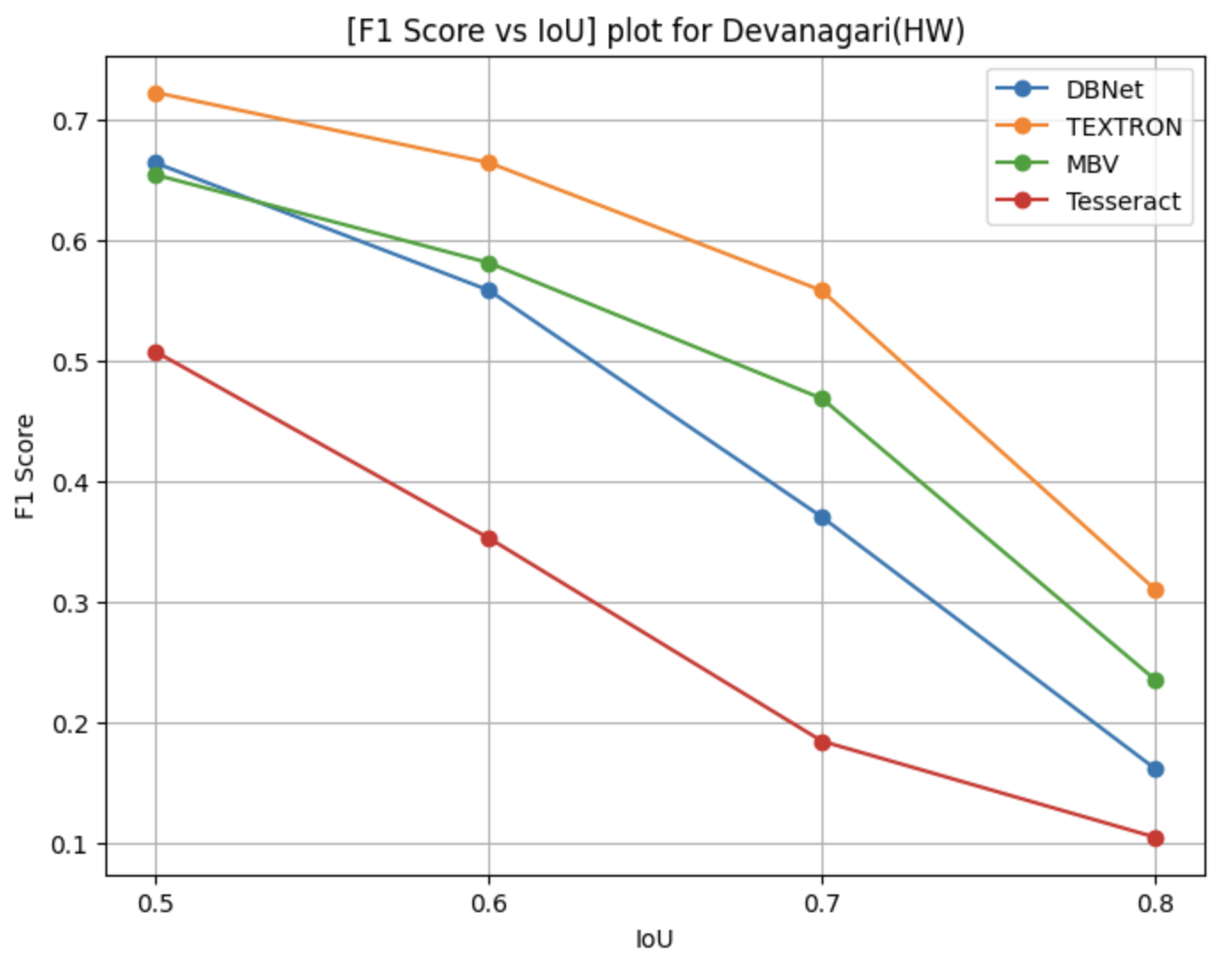}
    \caption{\Textron{ } performance against other baselines with  IOU thresholds on Handwritten Devanagari Text Detection} 
    \label{fig:graph}
\end{figure}

In this section, we evaluate and present the performance of $\Textron_{8LF}$ on different test sets. Table \ref{tab:1} presents a comparison of $\Textron_{8LF}$ along with other baselines for an IOU threshold of 0.5 on the Docbank test set described in Section \ref{dataset-Docbank}. We also highlight the performance of the annotated Indian language text detection datasets described in Sections \ref{dataset-indian}. We have used $\Textron_{8LF}$ configuration for carrying out the inference on these pages. We report our performance on all four language test sets along with other baselines in Table \ref{tab:indian-perf}. We can see that our approach has the highest recall for Malayalam text detection and is at par with the best-performing baseline. On the other hand, $\Textron_{8LF}$ gives the best overall performance on both Tamil and Gujarati text detection. The improvement brought about by $\Textron_{8LF}$ is visible in Figure \ref{fig:error-resolution}. Further, we also present our handwritten Devanagari text detection performance in Table \ref{tab:indian-perf}. The improvement in case of $\Textron_{8LF}$ for this handwritten text detection is substantially high. This is also visually depicted in Figures \ref{fig:err-dev} and \ref{fig:res-dev} respectively. Figure \ref{fig:graph} shows that $\Textron_{8LF}$ is able to maintain the best performance for Devanagari text detection even for higher IOU thresholds. We also highlight insights on the performances for other IOU scenarios and different classes of text (presented in the supplementary). This is beneficial for several downstream applications like handwritten text recognition.

\section{Conclusion and Future Work}

We present \Textron{}, a weak supervision-based approach that provides efficient results for text detection in multilingual documents. We demonstrate the flexibility of \Textron{ } by enabling the use of customized LFs to enhance text detection. The proposed approach is better off visually in addressing handwritten text and also near to state-of-the-art in other printed modalities. Our results also highlight that \Textron{ } can effectively be used to perform text detection in Indian language printed and handwritten documents. This also indicates its versatility by performing well even in the absence or paucity of labeled data. Our work paves the way in the direction of improving script-specific or font-specific text detection in documents. Plug and play with a set of customized LFs, setting intuitive quality guides, and tweaking hyperparameters can bring about enhanced text detection in different types of documents. Further work includes incorporating the detection of document classes other than text such as figures, tables, and equations. It will also be beneficial to expand the scope of our framework to detect text ranging from different kinds of handwritten documents to natural scenes, by designing suitable labeling functions.

\subsubsection*{Acknowledgements}
We acknowledge the support of a grant from IRCC, IIT Bombay, and
MEITY, Government of India, through the National Language Translation Mission-Bhashini project.

%%%%%%%%% REFERENCES
{\small
\bibliographystyle{ieee_fullname}
\bibliography{wacvbib}
}

\end{document}